\documentclass[journal]{IEEEtran}
\usepackage{subfigure}
\usepackage{multirow}
\usepackage{amsmath}
\usepackage{pifont}
\usepackage{hyperref}
\usepackage[pdftex]{graphicx}
\usepackage{cite}
\usepackage{verbatim}
\usepackage{color}
\newcommand{\cmark}{\ding{51}}   
\newcommand{\xmark}{\ding{55}}
\newcommand{\eg}{\textit{e.g. }}   
\newcommand{\ie}{\textit{i.e. }}
\begin{document}

\title{Is Heuristic Sampling Necessary in Training Deep Object Detectors?}

\author{Joya~Chen,~\IEEEmembership{Student~Member,~IEEE,}
  ~Dong~Liu,~\IEEEmembership{Senior~Member,~IEEE,}
  ~Tong~Xu,~\IEEEmembership{Member,~IEEE,}
  ~Shiwei~Wu, ~Yifei Cheng,
  and~Enhong~Chen,~\IEEEmembership{Senior~Member,~IEEE}
  \thanks{This work was partially supported by the grants from the National Key Research and Development Program of China under Grant 2018YFB1402600, and the National Natural Science Foundation of China under Contract 61727809 and 62072423. It was also supported by the GPU cluster built by Anhui Province Key Lab of Big Data Analysis and Application, University of Science and Technology of China (Corresponding author: Enhong Chen).}
  \thanks{Joya Chen, Tong Xu, Shiwei Wu, Yifei Cheng, and Enhong Chen are with Anhui Province Key Lab of Big Data Analysis and Application, University of Science and Technology of China. Among them, Joya Chen, Tong Xu, and Enhong Chen are with School of Computer Science and Technology; Shiwei Wu and Yifei Cheng are with School of Data Science (e-mail: \{chenjoya, dwustc, chengyif\}@mail.ustc.edu.cn, \{tongxu, cheneh\}@ustc.edu.cn).}
  \thanks{Dong Liu is with Department of Electronic Engineering and Information Science, University of Science and Technology of China (e-mail: dongeliu@ustc.edu.cn).}
}

\markboth{IEEE TRANSACTIONS ON IMAGE PROCESSING}
{Chen \MakeLowercase{\textit{et al.}}: Is Heuristic Sampling Necessary in Training Deep Object Detectors?}
\maketitle

\begin{abstract}
  To train accurate deep object detectors under the extreme foreground-background imbalance, heuristic sampling methods are always necessary, which either re-sample a subset of all training samples (hard sampling methods, \eg biased sampling, OHEM), or use all training samples but re-weight them discriminatively (soft sampling methods, \eg Focal Loss, GHM). In this paper, we challenge the necessity of such hard/soft sampling methods for training accurate deep object detectors. While previous studies have shown that training detectors without heuristic sampling methods would significantly degrade accuracy, we reveal that this degradation comes from an unreasonable classification gradient magnitude caused by the imbalance, rather than a lack of re-sampling/re-weighting. Motivated by our discovery, we propose a simple yet effective \emph{Sampling-Free} mechanism to achieve a reasonable classification gradient magnitude by initialization and loss scaling. Unlike heuristic sampling methods with multiple hyperparameters, our Sampling-Free mechanism is fully data diagnostic, without laborious hyperparameters searching. We verify the effectiveness of our method in training anchor-based and anchor-free object detectors, where our method always achieves higher detection accuracy than heuristic sampling methods on COCO and PASCAL VOC datasets. Our Sampling-Free mechanism provides a new perspective to address the foreground-background imbalance. Our code is released at \url{https://github.com/ChenJoya/sampling-free}.
\end{abstract}

\begin{IEEEkeywords}
  Object Detection, Foreground-Background Imbalance, Heuristic Sampling, Sampling-Free
\end{IEEEkeywords}

\section{Introduction}

\IEEEPARstart{W}{ith} the development of deep learning~\cite{alexnet,dl}, recent years have witnessed remarkable advancement in object detection~\cite{generic_od}. Among them, representative successes include two-stage R-CNN detectors~\cite{faster_rcnn,rfcn,fpn,mask_rcnn,relod,cascade_rcnn,iounet,grid_rcnn,libra_rcnn,htc,garpn,tridentnet}: their first stage uses a region proposal network (RPN~\cite{faster_rcnn}) to generate some candidates from dense, predefined bounding-boxes (\textit{i.e.} anchors), then the second stage uses a region-of-interest subnetwork (RoI-subnet) for object classification and localization. To pursue higher efficiency, one-stage approaches~\cite{ron,focal_loss,yolov2,yolov3,refinedet,rfbnet,fsaf,freeanchor} directly recognize objects from dense anchors rather than generating candidate proposals. Both two-stage and one-stage detectors adopt the anchoring scheme, where massive anchors ($\sim$$10^5$) are uniformly sampled over an image.

Nevertheless, when training these anchor-based detectors, only a few anchors ($\sim$$10^2$) that highly overlap with objects will be assigned to foreground samples, which always results in an extreme imbalance between foreground and background (\ie \textit{fg-bg} imbalance) within the anchors. In previous studies~\cite{focal_loss,ghm}, such imbalance may impede the training from convergence, as well as limit the detection accuracy. More recently, anchor-free object detectors~\cite{cornernet,extremenet,grid_rcnn,centernet_triplets,reppoints,dense_reppoints,fcos,garpn,foveabox,sapd} have gained much attention due to the replacement of anchors by points (\eg corner/center points), but they still suffer from the similar imbalance within the points.

To address the \textit{fg-bg} imbalance, several heuristic methods have been proposed to train deep object detectors in recent years. These methods can be divided into two categories. The first category re-samples a subset of training samples, \eg biased sampling~\cite{faster_rcnn}, online hard example mining~\cite{ohem} (OHEM), IoU-balanced sampling~\cite{libra_rcnn}. The second category re-weights training samples discriminatively, \ie assigns different weights to different training samples, like Focal Loss~\cite{focal_loss}, gradient harmonizing mechanism~\cite{ghm} (GHM), prime sample attention mechanism~\cite{pisa} (PISA). According to \cite{imbalance_review}, these two categories can be named ``hard sampling methods'' and ``soft sampling methods,'' respectively. We also use the term ``heuristic sampling methods'' to refer to them in the following. 

Although deep object detectors are always equipped with heuristic sampling methods, it is still very difficult to design a suitable hard/soft sampling strategy. Each heuristic sampling methods have a different re-sampling/re-weighting method --- as it is unknown which sample and what weighting value is better. For example, in GHM~\cite{ghm}, the authors hold the opinion that the optimal distribution of gradient is hard to define and requires further research. Moreover, heuristic sampling methods always introduce multiple hyperparameters, which requires laborious searching.

\begin{figure*}
  \centering
  \includegraphics[width=\linewidth]{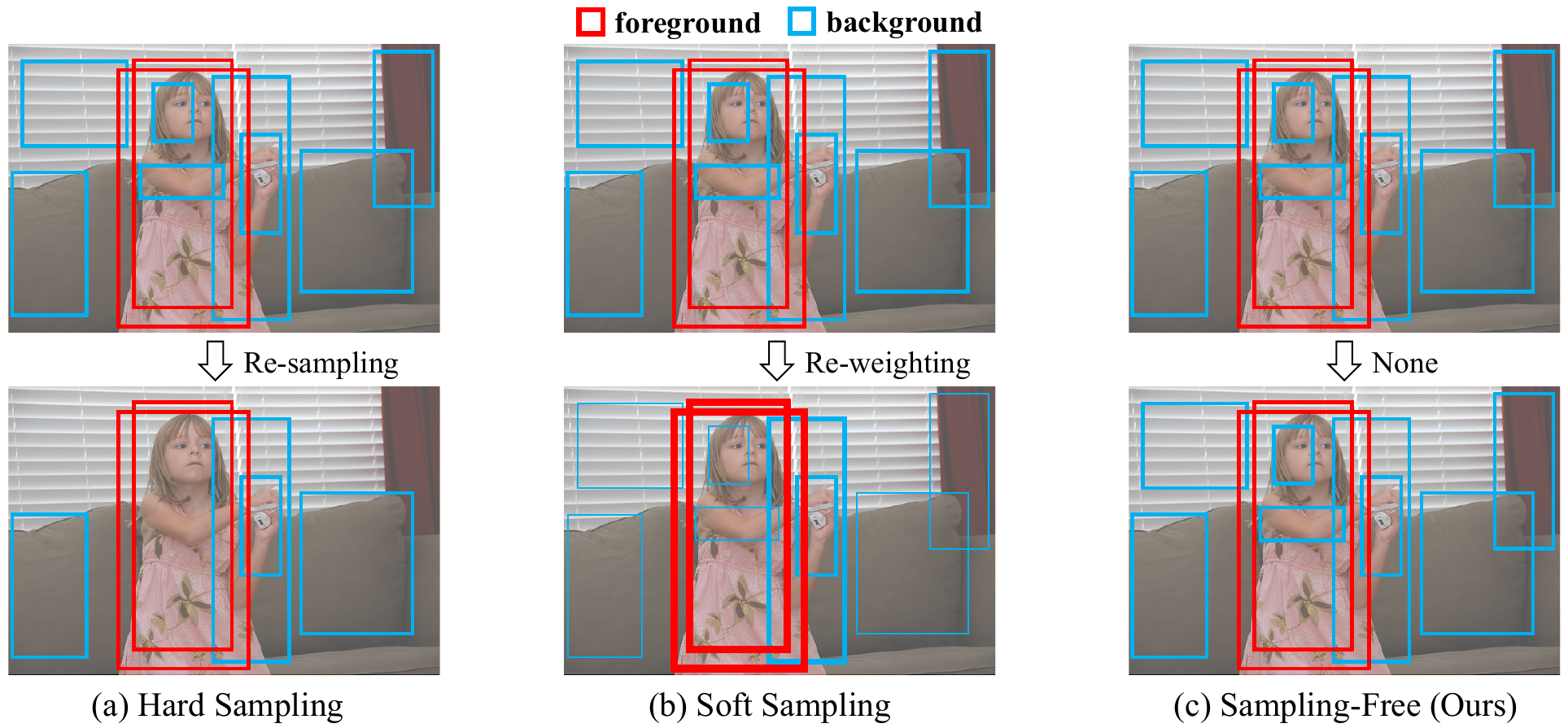}
  \caption{This figure illustrates the differences between heuristic sampling methods and the Sampling-Free mechanism in the treatment of training samples. Here we use the bounding-boxes to denote training samples (\eg anchors) in object detection. (a) Hard sampling (\eg biased sampling~\cite{faster_rcnn}, OHEM~\cite{ohem}, IoU-balanced sampling~\cite{libra_rcnn}) re-samples a subset of training samples; (b) Soft sampling (\eg Focal Loss~\cite{focal_loss}, GHM~\cite{ghm}, PISA~\cite{pisa}) uses all training samples but focuses on some of them by re-weighting. For instance, thicker boxes in (b) denote training samples with higher weights. (c) Sampling-Free equally uses all training samples.}
  \label{figure1}
\end{figure*}

Can we discard heuristic sampling methods when training deep object detectors? In the past, it was demonstrated~\cite{focal_loss,ghm} that the detector without heuristic sampling methods will suffer from the extreme \textit{fg-bg} imbalance, which would trail the detector with heuristic sampling methods about 20\% detection accuracy. Some methods~\cite{noisy, freeanchor, atss, paa, sapd, mal} adaptively define foreground/background labels to anchors/points, but they still rely on heuristic sampling methods to address the \textit{fg-bg} imbalance. Other ranking-based methods~\cite{ap_loss, dr_loss,alrp} try to avoid the \textit{fg-bg} imbalance by transforming the classification task into the ranking task, but they select pairs of $N$ samples to train, thus have much greater computational cost than heuristic sampling methods ($\mathcal{O}(N^2)$ vs. $\mathcal{O}(N)$). It seems impossible to make a cost-free replacement of heuristic sampling methods when training a deep object detector.

In this paper, we discover that a reasonable classification gradient magnitude is the key to address the \textit{fg-bg} imbalance, rather than hard/soft sampling. Motivated by this, we propose a simple yet effective \emph{Sampling-Free} mechanism that adaptively controls the classification gradient magnitude by initialization and loss scaling techniques, which enables discarding heuristic sampling methods but achieves better accuracy. Specifically, at the start of the training, the optimal bias initialization is used to reduce the excessive classification gradient magnitude caused by \textit{fg-bg} imbalance. During the training process, we leverage the bounding-box regression loss to adjust the classification loss, to achieve an adaptive adjustment for the classification gradient magnitude. As shown in Fig.~\ref{figure1}, unlike heuristic sampling methods, our method treats all training samples equally, without any hyperparameters introduced. 

Experimental results on COCO \cite{coco} and PASCAL VOC \cite{pascal_voc} datasets have demonstrated that our method is effective for both anchor-based and anchor-free object detectors, which always achieves higher detection accuracy than heuristic sampling methods. By replacing Focal Loss~\cite{focal_loss} with Sampling-Free in the adaptive label assignment strategy~\cite{paa}, we obtain a state-of-the-art 49.6 AP on COCO \texttt{test-dev}, without bells and whistles. Sampling-Free is also generalized for the instance segmentation task, which helps Mask R-CNN to obtain better segmentation accuracy. Moreover, no hyperparameter is introduced in our method. Our Sampling-Free mechanism provides a new perspective to address the \textit{fg-bg} imbalance.

Our main contributions are as follows: 

$\bullet$ For the first time, we discover what prevents detectors without heuristic sampling methods from achieving good accuracy --- the unreasonable classification gradient magnitude under the \textit{fg-bg} imbalance, rather than the lack of re-sampling/re-weighting on training samples.

$\bullet$ We propose a novel Sampling-Free mechanism that enables training deep object detectors without heuristic sampling methods. It adaptively controls the classification gradient magnitude by initialization and loss scaling, which is easy to implement and introduces no hyperparameters.

$\bullet$ Collaborating with Sampling-Free mechanism, it is feasible to train deep object detectors without any hard/soft sampling methods and achieve better results on COCO and PASCAL VOC benchmarks.

\section{Related Work}

Classical object detectors~\cite{dpm,jones} usually rely on hand-crafted feature extractor, which is hard to design. With the development of deep learning~\cite{alexnet,dl}, deep object detectors quickly come to dominate object detection. In this section, we introduce the development of deep object detectors, then introduce the concept, cause, and solution of the \textit{fg-bg} imbalance. Finally, we discuss the relations and differences between our work and previous works.

\subsection{Deep Object Detection}

Among deep object detectors, the anchor-based approach is the most popular approach, which tiles massive default bounding-boxes (\ie anchors) on an image to cover objects. There are mainly two types of anchor-based approaches: 

\textit{1) Two-stage Anchor-based:} It is popularized by Faster R-CNN~\cite{faster_rcnn}, which firstly generates some candidates from massive anchors by region proposal network (RPN~\cite{faster_rcnn}), then determines the accurate location and object category by a subnetwork (RoI-subnet~\cite{rcnn,fast_rcnn}). A large number of Faster R-CNN variants~\cite{rfcn,fpn,mask_rcnn,relod,cascade_rcnn,iounet,grid_rcnn,libra_rcnn,htc,garpn,tridentnet} appear over the years, yielding a large improvement in detection accuracy. 

\textit{2) One-stage Anchor-based:} It is popularized by SSD~\cite{ssd}, which performs much faster than two-stage detectors due to the elimination of the RPN, but usually achieves lower accuracy than two-stage detectors. A series of advances~\cite{ron,focal_loss,yolov2,yolov3,refinedet,rfbnet,fsaf,freeanchor, mal} in recent years promote one-stage anchor-based detector to be more accurate.

In recent years, a large number of anchor-free approaches are proposed, which detect objects by points or regions rather than anchors. Most anchor-free object detectors follow the one-stage detection pipeline but eliminate the usage and hyperparameters of ``anchor boxes'', which shows better simplicity. Although the differences among anchor-free detectors are much smaller than the differences between one-stage and two-stage anchor-based detectors, there are some subtle differences in the definition of training samples among different anchor-free detectors. Specifically, some of the anchor-free approaches detect objects by generating bounding-boxes according to pre-defined or self-learned keypoints. The early attempt of pre-defined points is CornerNet~\cite{cornernet}, which uses the top-left corner and bottom-right corner to represent objects. After that, researchers use various pre-defined points to represent objects, such as extreme points~\cite{extremenet}, grid points~\cite{grid_rcnn}, center points~\cite{centernet_triplets}, and self-learned points~\cite{reppoints,dense_reppoints}. Others~\cite{fcos,foveabox, sapd,garpn} try to learn the position of the object center, and then regress the distances from the center to the four sides of the object bounding-box for detection. The most popular center-based anchor-free detector is FCOS~\cite{fcos}, which regards all the locations around the center of objects as foreground examples. GA-RPN~\cite{garpn} successfully designs an anchor-free RPN in the two-stage pipeline. 

\subsection{Foreground-Background Imbalance Problem}

Training a deep object detector involves two tasks: classification and localization. For classification, the number of background examples is much larger than foreground examples, which is known as the \textit{fg-bg} imbalance~\cite{focal_loss}. We introduce it in the different label assignment strategy:

\textit{1) Fg-bg Imbalance in Fixed Label Assignment:} In the fixed label assignment~\cite{faster_rcnn,ssd,cornernet,fcos}, there is a pre-defined rule to assign a training sample to a \textit{fg}/\textit{bg} example. For instance, the anchor-based approach usually considers anchors that have large intersection-over-union (IoU) with ground-truths as foreground examples (\eg IoU$>$0.5). The anchor-free approach usually regards points around the center as foreground examples. However, the total number of anchors/points is always huge, which may be $10^3$$\sim$$10^4$ times more than foreground anchors/points. Therefore, during training, the \textit{fg-bg} imbalance inevitably occurs in the classification task.

\textit{2) Fg-bg Imbalance in Adaptive Label Assignment:} Recently, several adaptive label assignment methods~\cite{noisy, freeanchor, atss, paa, sapd, mal} are proposed to overcome the limitations of fixed label assignment. However, they still suffer from extreme \textit{fg-bg} imbalance. For example, FreeAnchor~\cite{freeanchor} claims that it faces an even more serious sample imbalance than RetinaNet~\cite{focal_loss}. These methods still rely on Focal Loss~\cite{focal_loss} to address the \textit{fg-bg} imbalance in the classification task.

\subsection{Solutions for Foreground-Background Imbalance}

As we can see, the \textit{fg-bg} imbalance always exists in training deep object detectors, which impedes deep object detectors from achieving higher accuracy as reported in~\cite{focal_loss,ghm}. In previous works, there are three solutions:

\textit{1) Sampling Methods:} It is the most common solution for \textit{fg-bg} imbalance, which has two groups~\cite{imbalance_review} --- hard sampling and soft sampling. The hard sampling method re-samples a set of training samples by some strategies. For example, biased sampling~\cite{faster_rcnn} randomly samples 256 examples with 1:1 \textit{fg-to-bg} ratio during training RPN. OHEM~\cite{ohem} and IoU-balanced sampling~\cite{libra_rcnn} selects training samples according to loss and IoU values, respectively. The objectness~\cite{ron,yolov3,refinedet} modules, generative methods~\cite{proi, a_fast_rcnn} can also be regarded as hard sampling methods. Soft sampling re-weights training samples discriminatively by some strategies. Focal Loss~\cite{focal_loss} dynamically assigns higher weight to the hard training samples (\ie with high loss value). Similar to Focal Loss, GHM~\cite{ghm} suppresses gradients originating from easy and very hard training samples (\ie with low loss value). PISA~\cite{pisa} re-weights training samples according to the IoU between training samples and ground-truths.

\textit{2) Classification to Ranking:} As the \textit{fg-bg} imbalance always exists in classification task in deep object detectors, AP Loss~\cite{ap_loss} and DR Loss~\cite{dr_loss} propose to convert the classification task into ranking task. These methods train a pair of samples rather than independent sample. Specifically, the predicted score of one training sample is transformed into the difference between the predicted scores of two training samples. These methods are also quite in line with the detection evaluation metric (average precision, AP).

To date, almost all deep object detectors are equipped with sampling methods during training. The ranking task, however, trains pairs of $N$ examples thus has $\mathcal{O}(N^2)$ computational cost, which is much higher than $\mathcal{O}(N)$ cost of the classification task. Although sampling methods are popular, the re-sampling/re-weighting strategy is hard to design, and both sampling methods and ranking-based methods require laborious hyperparameter tuning. Our work overcomes these shortcomings, which discards sampling methods in the classification task without any hyperparameters introduced.

\section{Investigation}

As shown in Table~\ref{table1}, it was believed~\cite{focal_loss,ghm} that under the \textit{fg-bg} imbalance, training a deep object detector without heuristic sampling methods will lead to a nearly $20\%$ decrease in the detection accuracy. In this section, we will investigate why this is the case. It will be revealed that the decrease should be attributed to the unreasonable gradient magnitude, rather than the lack of re-sampling/re-weighting on training samples. 

\begin{table}[t]
  \centering
  \caption{Comparison of different classification loss functions in object detection. This table comes from ``Gradient Harmonized Single-Stage Detector''~\cite{ghm}.}
  \begin{tabular}[t]{|c|cccccc|}
    \hline
    method & AP & AP$_{50}$ & AP$_{75}$ & AP$_{S}$ & AP$_{M}$ & AP$_{L}$   \\ 
    \hline
    CE loss & 28.6 & 43.3 & 30.7 & 11.4 & 30.7 & 40.7 \\
    OHEM~\cite{ohem} & 31.1 & 47.2 & 33.2 & - & - & - \\
    Focal Loss~\cite{focal_loss} & 35.6 & 55.6 & 38.2 & 19.1 & 39.2 & 46.3 \\
    GHM-C~\cite{ghm} & 35.8 & 55.5 & 38.1 & 19.6 & 39.6 & 46.7 \\
    \hline
  \end{tabular}
  \label{table1}
\end{table}

In the following, we first mathematically introduce \textit{fg-bg} imbalance and sampling methods, then theoretically analyze their impact on the training process. Finally, we experimentally demonstrate that the gradient magnitude is the key factor affecting the detector accuracy. 

\subsection{Concepts}\label{section3.1}

\textit{1) Foreground-Background Imbalance:} In general, deep object detectors tend to generate numerous samples to cover as many objects as possible. Although there are various label assignment strategies~\cite{faster_rcnn, focal_loss, freeanchor, atss, paa} to define foreground and background samples, the imbalance between foreground and background samples will be inevitably caused due to the rarity of objects and the majority of samples, namely foreground-background (\textit{fg-bg}) imbalance. In other words, the number of foreground samples $N^f$ is much smaller than that of background samples $N^b$ (\ie $N^f \ll N^b$). Unlike the common class imbalance caused by the biased dataset, the \textit{fg-bg} imbalance is more likely to be introduced by the ``numerous'' sample generation strategy. Thus, for a deep object detector, the \textit{fg-bg} imbalance is similarly distributed during training and inference, as the detector always shares the same sample generation strategy in those two phases. 

\textit{2) Heuristic Sampling Methods:} Whether hard or soft sampling, the essence is to re-sample or re-weight training samples in the loss computation. If we regard hard sampling as soft sampling with weights 0 or 1, both of them be summarized as 

\begin{eqnarray} 
  L = s\sum_{i}^{N}w_il_i, \label{equation1}
\end{eqnarray}

\noindent where $L$ denotes the overall training loss for a batch, and $s$ is a scaling term. $w_i$ and $l_i$ are the weight and the loss of $i$-th sample in a batch, respectively. In general, deep object detectors uses cross-entropy (CE) loss as $l_i$ in the classification task, \ie

\begin{eqnarray} 
  L = -s\sum_{i}^{N}w_i[y_i\log(p_i) + (1 - y_i)\log(1-p_i)], \label{equation2}
\end{eqnarray}

\noindent where $y_i$ is the ground-truth label for $i$-th sample: $y_i = 1$ if it is foreground, otherwise $y_i = 0$. $p_i$ is the confidence score, ranging from 0 to 1. Since the weights for foreground and background samples usually have different forms, Equation~\ref{equation2} can be further rewritten as

\begin{eqnarray} 
  L = -s\sum_{i}^{N}y_iw_i^f\log(p_i) + (1 - y_i)w_i^b\log(1-p_i), \label{equation3}
\end{eqnarray}

\noindent where the notations $f$ and $b$ denote foreground and background, respectively. Equation~\ref{equation3} can also describe the training loss without sampling methods. In that case, we have $w_i^f = w_i^b = 1$ for all training samples.

\subsection{Analysis}\label{section3.2}

As most deep object detectors are trained with mini-batch stochastic gradient descent (mini-batch SGD), we discuss here the effect of the \textit{fg-bg} imbalance and heuristic sampling methods on mini-batch SGD training. For each iteration, the learnable parameters $\Theta$ of the detector will be updated in the direction of the gradient, \ie

\begin{eqnarray} 
  \Theta^{t+1}  = \Theta^t - \eta\frac{\partial L}{\partial \Theta^t}, \label{equation4}
\end{eqnarray}

\noindent where $\Theta^{t}$ denotes the parameters in $t$-th step, and $\eta$ is the learning rate. According to Equation~\ref{equation3}, the gradient can be further expressed as

\begin{eqnarray} 
  \frac{\partial L}{\partial \Theta^t} &=& -s\sum_{i}^{N}y_i\frac{\partial [w_i^f\log(p_i)]}{\partial \Theta^t} \nonumber \\ &+& (1 - y_i)\frac{\partial [w_i^b\log(1-p_i)]}{\partial \Theta^t}. \label{equation5}
\end{eqnarray}

\noindent As the exact form of the parameters $\Theta^t$ is unknown, a quantitative analysis for mini-batch SGD training seems impossible, especially after multiple training iterations. Therefore, in the following, we turn to analyze the case at the start of the training. We denote the learning parameters at the start of the training as $\Theta_s$. At this point, $\Theta_s$ cannot distinguish foreground samples from background samples. In other words, the detector outputs similar confidence scores $p_i \approx p$ for all samples. Then, we have

\begin{eqnarray} 
  \frac{\partial L}{\partial \Theta^s} &\approx& -s\sum_{i}^{N}y_i\frac{\partial [w_i^f\log(p)]}{\partial \Theta^s} \nonumber \\ &+& (1 - y_i)\frac{\partial [w_i^b\log(1-p)]}{\partial \Theta^s} \nonumber \\ &\approx& -s\sum_{i}^{N}y_i[\frac{\partial w_i^f}{\partial \Theta^s}\log(p) + \frac{w_i^f}{p}\frac{\partial p}{\partial \Theta^s}] \nonumber \\ &+& (1 - y_i)[\frac{\partial w_i^b}{\partial \Theta^s}\log(1-p) - \frac{w_i^b}{1 - p}\frac{\partial p}{\partial \Theta^s}]. \label{equation6}
\end{eqnarray}

\noindent If the weights $w_i^f$ and $w_i^b$ are constants, Equation~\ref{equation6} would be simple as $\frac{\partial w_i^f}{\partial \Theta^s} = \frac{\partial w_i^b}{\partial \Theta^s} = 0$. However, in most soft sampling methods~\cite{focal_loss,ghm,pisa}, the weight of a training sample is usually dynamic that may depend on confidence scores, IoU scores, and training iterations. Therefore, we would like to discuss the training cases of constant weights and dynamic weights, respectively. 

\textit{1) Constant Weights:} In this case, we have $w_i^f = w^f, w_i^b = w^b$ for any $i$, where $w^f$ and $w^b$ are constants. Thus, we also have $\frac{\partial w_i^f}{\partial \Theta^s} = \frac{\partial w_i^b}{\partial \Theta^s} = 0$, and Equation~\ref{equation6} can be derived as

\begin{eqnarray} 
  \frac{\partial L}{\partial \Theta^s} &\approx& -s\sum_{i}^{N}y_i\frac{w^f}{p}\frac{\partial p}{\partial \Theta^s} - (1 - y_i)\frac{w^b}{1 - p}\frac{\partial p}{\partial \Theta^s} \nonumber \\ &=& -s(N^f\frac{w^f}{p} - N^b\frac{w^b}{1 - p})\frac{\partial p}{\partial \Theta^s}, \label{equation7}
\end{eqnarray}

\noindent where $N^f$ and $N^b$ are the number of foreground and background training samples in a training iteration, respectively. Following \cite{convex}, we use $||\cdot||$ to denote the L2-norm of a gradient vector, which represent its magnitude. From Equation~\ref{equation7}, the gradient magnitude is

\begin{eqnarray} 
  ||\frac{\partial L}{\partial \Theta^s}|| \approx s|N^f\frac{w^f}{p} - N^b\frac{w^b}{1 - p}|\cdot||\frac{\partial p}{\partial \Theta^s}||. \label{equation8}
\end{eqnarray}

If $\Theta_s$ is not biased, the initial estimate for both foreground and background samples are $p = 0.5$, then we have

\begin{eqnarray} 
  ||\frac{\partial L}{\partial \Theta_s}(p=0.5)|| \approx 2s |N^fw^f - N^bw^b| \cdot ||\frac{\partial p}{\partial \Theta^s}||. \label{equation9}
\end{eqnarray}

\noindent When heuristic sampling methods are not used, \ie $w^f_i = w^b_i = 1$, the gradient magnitude of the \textit{fg-bg} imbalance case ($N_f \ll N_b$) will be much larger than that of the balanced case ($N_f \approx N_b$). Thus, if we train a detector without heuristic sampling methods, the \textit{fg-bg} imbalance will result in a much larger gradient magnitude at the start of the training. If the scaling term $s$ is not set properly, it may lead to training divergence. Compared with the gradient magnitude with heuristic sampling methods, the weighting terms $w^f$ and $w^b$ can alleviate the imbalance between $N_f$ and $N_b$, thus leading to better stability at the start of the training.

\textit{2) Dynamic Weights:} In this case, we take the well-known Focal Loss~\cite{focal_loss} as an example, which proposes a unified representation of the weighting term as $\alpha_t(1-p_t)^\gamma$. When we apply it separately for foreground and background samples, we have $w_i^f = \alpha(1 - p_i)^\gamma$ and $w_i^b = (1 -\alpha)p_i^\gamma$. Here $\alpha$ and $\gamma$ are the hyperparameters in Focal Loss for adaptively re-weighting training samples. As $p_i \approx p$ at the start of the training, we have $w_i^f \approx \alpha(1 - p)^\gamma$ and $w_i^b \approx (1 -\alpha)p^\gamma$. Then, Equation~\ref{equation6} can be derived as

\begin{eqnarray} 
  \frac{\partial FL}{\partial \Theta^s} &\approx& -s\sum_{i}^{N}y_i\frac{\partial p}{\partial \Theta^s}[\frac{\partial w_i^f}{\partial p}\log(p) + \frac{w_i^f}{p}] \nonumber \\ &+& (1 - y_i)\frac{\partial p}{\partial \Theta^s}[\frac{\partial w_i^b}{\partial p}\log(1-p) - \frac{w_i^b}{1 - p}] \nonumber \\ &\approx& -s\frac{\partial p}{\partial \Theta^s}\sum_{i}^{N}y_i\alpha(1-p)^{\gamma-1}[-\gamma\log(p) + \frac{1-p}{p}] \nonumber \\ &+& (1 - y_i)(1-\alpha)p^{\gamma-1}[\gamma\log(1-p) - \frac{p}{1 - p}] \nonumber \\ &\approx& -s\frac{\partial p}{\partial \Theta^s}\{N^f\alpha(1-p)^{\gamma-1}[-\gamma\log(p) + \frac{1-p}{p}] \nonumber \\ &+& N^b(1-\alpha)p^{\gamma-1}[\gamma\log(1-p) - \frac{p}{1 - p}]\},  \label{equation10}
\end{eqnarray}

\noindent where $FL$ denotes Focal Loss. As reported in \cite{focal_loss}, the best setting of Focal Loss is $\alpha=0.25, \gamma=2$ on COCO dataset~\cite{coco}, and Focal Loss uses a biased initialization to ensure $p\approx0.01$. With these values, the gradient magnitude can be computed as

\begin{eqnarray} 
  ||\frac{\partial FL}{\partial \Theta^s}|| \approx 2s|10N^f - 10^{-4}N^b|\cdot||\frac{\partial p}{\partial \Theta^s}||.  \label{equation11}
\end{eqnarray}

\noindent As we can see, the RHS (right hand side) of Equation~\ref{equation11} is equal to the RHS of Equation~\ref{equation9} by setting $w^f=10, w^b=10^{-4}$. Coincidentally, for COCO~\cite{coco} dataset, a training anchor will learn 80 binary classifies for 80 object classes. In our observation, the \textit{fg-to-bg} ratio of training anchor is $1:10^3$, thus $\frac{N^b}{N^f} \approx 8 \times 10^4$, which is close to $\frac{w^f}{w^b}=10^5$. As we can see, Focal Loss also tries to alleviate the imbalance between $N_f$ and $N_b$, to obtain a reasonable gradient magnitude.

However, this does not mean that it is impossible to train the detector without heuristic sampling methods. In fact, we can reduce $s$ to lower the excessive gradient magnitude. But this in turn creates the dilemma of too small gradient magnitude. Specifically, when we set a small $s_m$ as the scaling factor to train a detector without heuristic sampling methods, as $N^f \ll N^b$, $p$ will rapidly approach $p \approx \frac{N^f}{N^f+N^b}$ to achieve a minimal loss value, and the gradient magnitude

\begin{eqnarray} 
  ||\frac{\partial L}{\partial \Theta_s}(w^f=w^b=1,s=s_m)|| \nonumber \\ \approx s_m |\frac{N^f}{p} - \frac{N^b}{1 - p}|\cdot||\frac{\partial p}{\partial \Theta^s}|| \label{equation12}
\end{eqnarray}

\noindent will be greatly decreased. At this point, $s_m$ becomes unreasonable and we should set a large scaling factor. It is not surprising why most effective heuristic sampling methods are dynamic, like Focal Loss~\cite{focal_loss} and GHM~\cite{ghm}. 

To sum up, it is essential to control the gradient magnitude in the classification task when training a detector without heuristic sampling methods. As illustrated in ~\cite{gradnorm,alrp}, the gradient magnitude will have a significant impact on the performance of multi-task learning. Object detection usually involves two or more tasks, the unreasonable gradient magnitude on the classification task affects not only itself but also other tasks. But in fact, we have various ways to control the gradient magnitude, and the heuristic sampling method is not the only choice.

Furthermore, the \textit{fg-bg} imbalance, as we illustrated in Sec.~\ref{section3.1}, also has the similar distribution in training and inference. If we use heuristic sampling methods during training, then it is equivalent to breaking the consistency of this distribution. In other words, $\Theta^e$ is obtained from the weighted imbalance distribution, which may not perform well in the vanilla imbalance distribution. Next, we will experimentally explore how to train deep object detectors without heuristic sampling methods. 

\begin{figure*}[t]
  \centering
  \subfigure[$\pi = 10^{-2}$ for \textit{RetinaNet-CE}]{ 
  \includegraphics[width=0.475\linewidth]{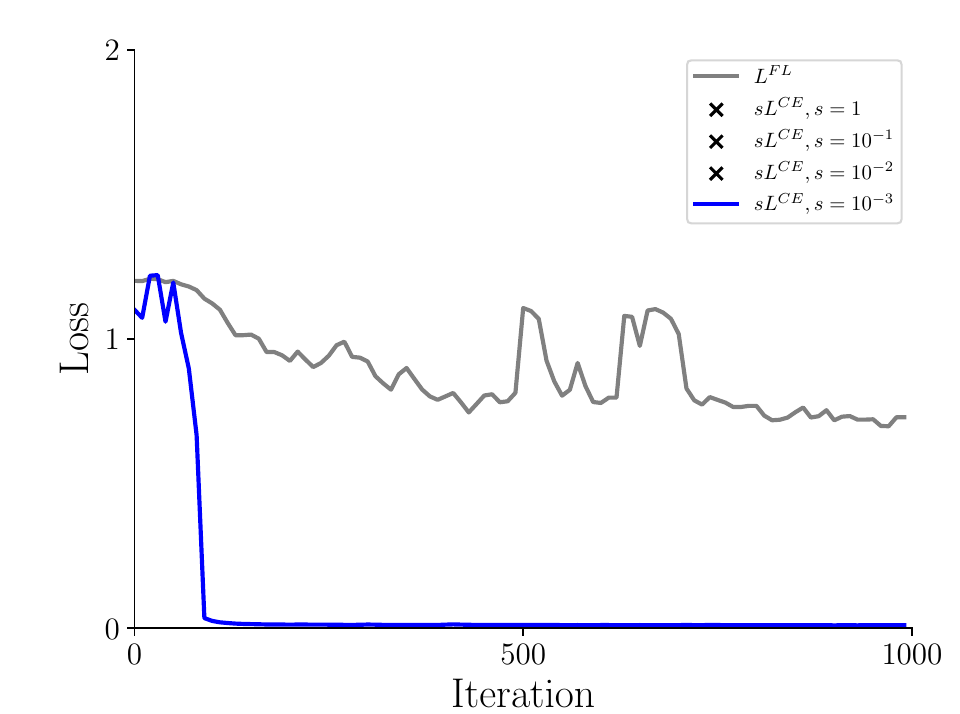} 
  }
  \subfigure[$\pi = 10^{-3}$ for \textit{RetinaNet-CE}]{ 
  \includegraphics[width=0.475\linewidth]{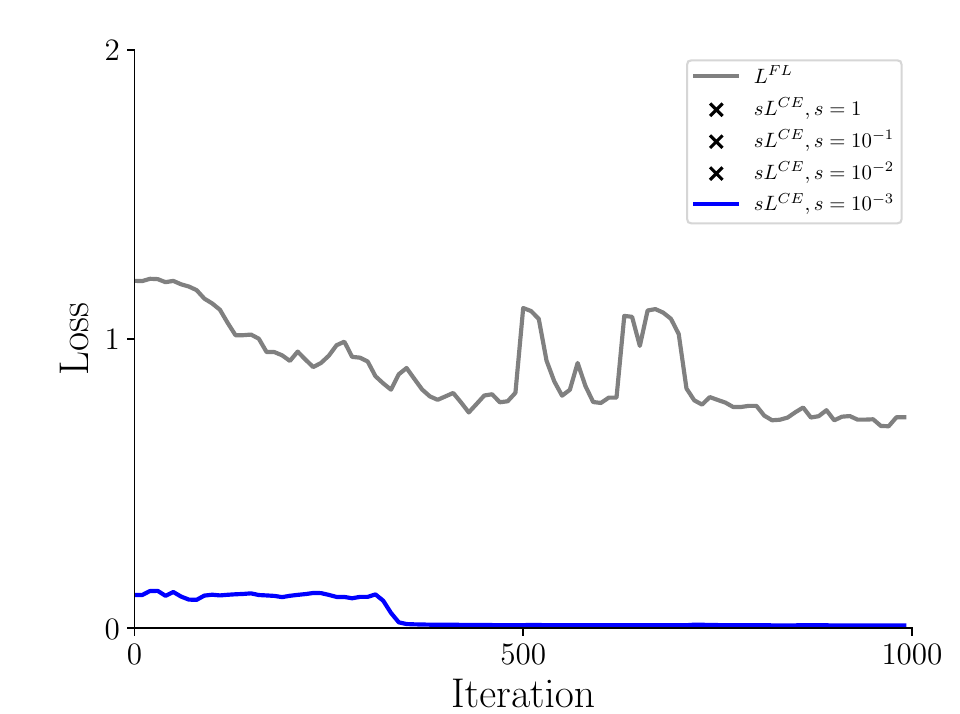} 
  }
  \subfigure[$\pi = 10^{-4}$ for \textit{RetinaNet-CE}]{
  \includegraphics[width=0.475\linewidth]{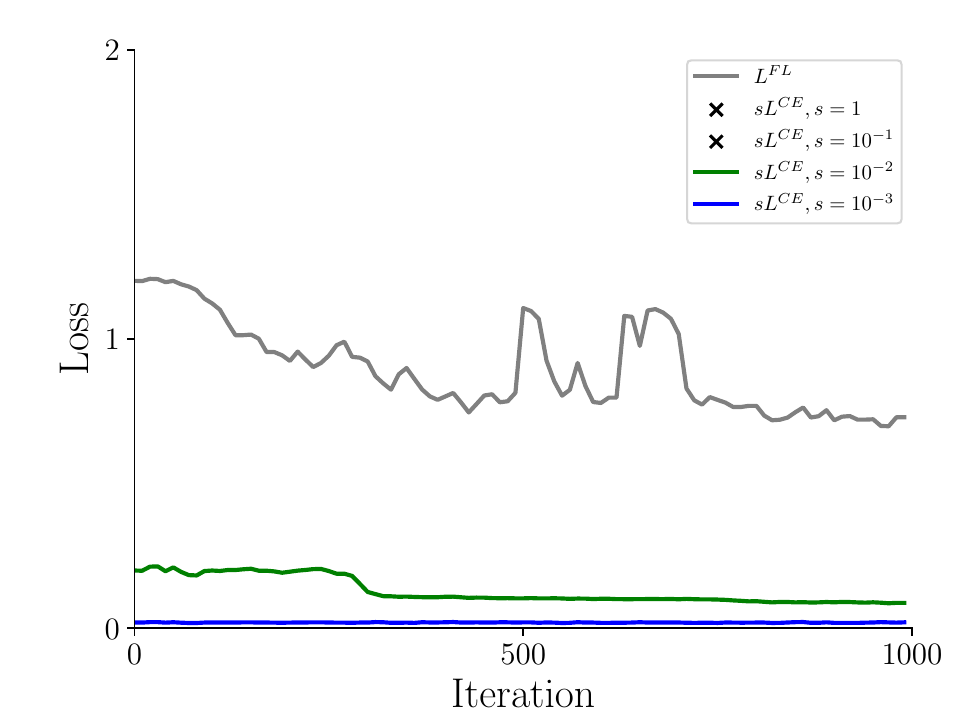} 
  }
  \subfigure[$\pi = 10^{-5}$ for \textit{RetinaNet-CE}]{
  \includegraphics[width=0.475\linewidth]{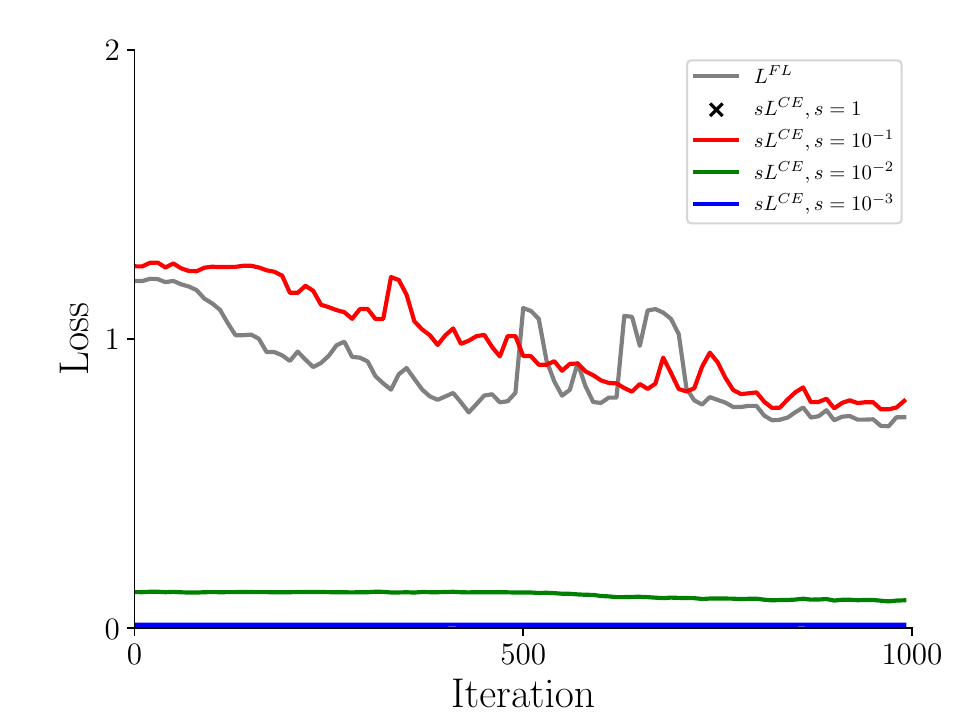} 
  }
  \caption{Loss curves of Focal Loss $L^{FL}$ and cross-entropy loss $L^{CE}$. ``\textbf{\xmark}'' means the network diverging. The detector is RetinaNet with ResNet-50-FPN~\cite{resnet,fpn} backbone, trained on COCO \texttt{train2017}~\cite{coco} with $1\times$ learning schedule~\cite{detectron} (12 epochs), implemented on \texttt{maskrcnn-benchmark}~\cite{maskrcnn_benchmark}. We only show the first $1k$ iterations for better visualization.}
  \label{figure2}
\end{figure*}

\subsection{Experimental Verification}\label{section3.3}

To investigate the accuracy gap between the detector with and without sampling methods, we experimentally investigate the difference between well-known Focal Loss~\cite{focal_loss} and CE loss. Focal Loss is widely used to address the \textit{fg-bg} imbalance in the one-stage anchor-based and anchor-free object detectors~\cite{focal_loss,fsaf,freeanchor,cornernet,extremenet,centernet_triplets,reppoints,dense_reppoints,fcos,garpn,foveabox,sapd, mal}. In previous studies~\cite{ghm,focal_loss}, Focal Loss helps RetinaNet~\cite{focal_loss} to yield 4$\sim$7 higher AP on COCO~\cite{coco} than CE loss. For simplicity, we denote the RetinaNet with Focal Loss and CE loss as \textit{RetinaNet-FL} and \textit{RetinaNet-CE}, respectively.

Two differences exist between \textit{RetinaNet-FL} and \textit{RetinaNet-CE}, one of which is the classification loss. Specifically, \textit{RetinaNet-FL} and \textit{RetinaNet-CE} use Focal Loss ($L^{FL}$) and CE loss ($L^{CE}$) in the classification task, respectively. Following the notations in Sec.~\ref{section3.1} and Sec.~\ref{section3.2}, we have

\begin{eqnarray}
  L^{FL}=\frac{1}{N^f}\sum_{i=1}^{N}w_il_i, \quad L^{CE} =\frac{1}{N^f}\sum_{i=1}^{N}l_i, \label{equation13}
\end{eqnarray} 

\noindent where $w_i$ is the weighting term, and $w_i = \alpha(1 - p_i)^\gamma$ if $i$-th training sample is foreground, otherwise $w_i = (1 -\alpha)p_i^\gamma$.

Another difference between \textit{RetinaNet-FL} and \textit{RetinaNet-CE} is the initialization. Focal Loss uses a biased initialization that initializes the final classification convolutional layer with the bias $b=-\log\frac{1-\pi}{\pi}$. Then, at the start of the training, $p_i=\pi$ is tenable for every training anchor as $p_i$ is computed by sigmoid activation. A heuristic value $\pi=10^{-2}$ is used in Focal Loss to avoid network diverging. Unfortunately, this will result in network diverging during training \textit{RetinaNet-CE}. We will also explore how to adjust $\pi$ in the following.

\begin{figure*}[t]
  \centering
  \subfigure[\textit{RetinaNet-FL}]{
  \includegraphics[width=0.475\linewidth]{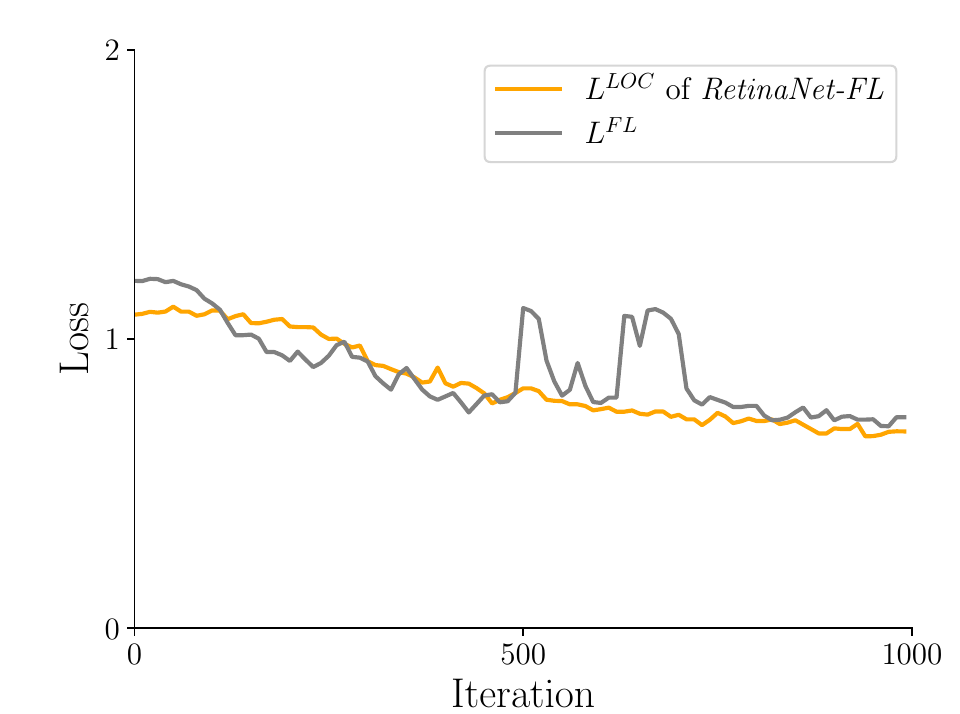} \label{figure3a}
  }
  \subfigure[\textit{RetinaNet-CE} ($\pi = 10^{-5}$)]{
  \includegraphics[width=0.475\linewidth]{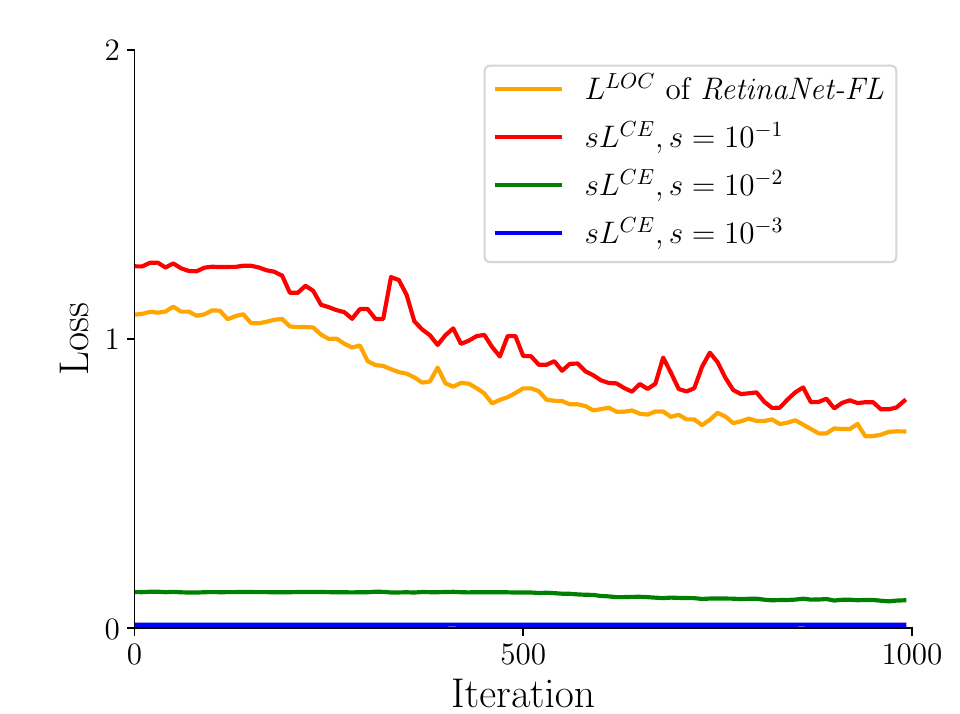} \label{figure3b}
  }
  \caption{Loss curves of localization ($L^{LOC}$) and classification ($L^{FL}$ or $L^{CE}$), which are from the same models in Fig.~\ref{figure2}. Note that the curves of $L^{LOC}$ with different $s$ in (b) are very similar, thus we only show the curve of $L^{LOC}$ with $s=10^{-1}$ in (b).}
  \label{figure3}
\end{figure*} 

\textit{1) Classification Loss:}
Our first discovery is that \textit{RetinaNet-CE} has poor stability on the classification loss, which reflects the unreasonable classification gradient magnitude. This phenomenon corresponds to our discovery in Sec.~\ref{section3.2}. Training \textit{RetinaNet-CE} with default $\pi=10^{-2}$ causes a large classification loss, with the network diverging in a few iterations. See Fig.~\ref{figure2}, we observe that \textit{RetinaNet-FL} can be stably trained, but training \textit{RetinaNet-CE} needs to carefully tune $\pi$ and $s$ to avoid the large classification loss. Only with specific settings, we can obtain several converged models, as shown in Fig.~\ref{figure2}. 

\begin{table}[t] 
\small
\centering
\caption{Detection accuracy of CE loss on COCO \texttt{minival}, with different $s$ and $\pi$. ``n/a'' refers to the network diverging.}
\subtable[Accuracy of \textit{RetinaNet-CE} in different $\pi, s$]{
\begin{tabular}[t]{|c|c|c|c|c|}
  \hline
  AP & $s = 1$ & $s = 10^{-1}$ & $s = 10^{-2}$ & $s = 10^{-3}$ \\ 
  \hline
  $\pi = 10^{-2}$ & n/a & n/a & n/a & 1.6  \\
  $\pi = 10^{-3}$ & n/a & n/a & n/a & 1.6  \\
  $\pi = 10^{-4}$ & n/a & n/a & 26.8 & 1.6  \\
  $\pi = 10^{-5}$ & n/a & \textbf{36.3} & 26.8 & 1.6   \\
  \hline
\end{tabular}\label{table2a}
}
\subtable[Accuracy comparison between \textit{RetinaNet-FL} and \textit{RetinaNet-CE}]{
\begin{tabular}[t]{|c|c|c|c|}
  \hline
  Model & AP & AP$_{50}$ & AP$_{75}$ \\
  \hline
  \textit{RetinaNet-FL} & \textbf{36.4} & \textbf{55.0} & \textbf{39.0}  \\
  \hline
  \textit{RetinaNet-CE} ($\pi=10^{-5}, s=10^{-1}$) & 36.3 & 54.8 & 38.7 \\
  \hline
\end{tabular}\label{table2b}
}
\end{table}

\textit{2) Detection Accuracy:}
Our second discovery is that the classification loss scale will greatly influence the accuracy of the \textit{RetinaNet-CE}. As shown in Table~\ref{table2a}, if a model can be stably trained, then its accuracy will be similar when $s$ is fixed, but tuning $s$ leads to great changes in AP. Table~\ref{table2b} shows that \textit{RetinaNet-CE} with $\pi=10^{-5}, s=10^{-1}$ has already achieved the comparable accuracy of \textit{RetinaNet-FL} (36.3 AP vs. 36.4 AP). This is an inspiring result, as previous works~\cite{ghm,focal_loss} reported there is a 4$\sim$7 AP gap between CE loss and Focal Loss. Our investigation successfully narrows this gap. However, it is still unknown why adjusting $\pi$ and $s$ can help to train \textit{RetinaNet-CE}. Next, we will perform an analysis of this.

\textit{3) Analysis:} 
At the start of the training, the bias initialization ensures $p_{i} \approx \pi$ for each training sample. According to Eq.~\ref{equation2} and Eq.~\ref{equation13}, we can estimate $L^{FL}$ as 

\begin{eqnarray}
  L^{FL} \approx -\alpha(1-\pi)^{s}\log(\pi) - (1 - \alpha)\pi^s\frac{N^b}{N^f}\log(1 - \pi),
\end{eqnarray}

\noindent and estimate $L^{CE}$ as

\begin{eqnarray}
  sL^{CE} \approx -s\log(\pi) - s\frac{N^b}{N^f}\log(1 - \pi).
\end{eqnarray}

For COCO~\cite{coco} dataset, a training anchor will learn 80 binary classifies for 80 object classes. In our observation, the \textit{fg-to-bg} ratio of training anchor is $1:10^3$, thus $\frac{N^b}{N^f} \approx 10^3 \times 80$. With $\alpha=0.25, s=2, \pi=10^{-2}$ in training \textit{RetinaNet-FL}, and $s=10^{-1},\pi=10^{-5}$ in training \textit{RetinaNet-CE}, we can estimate $L^{FL}$ and $L^{CE}$ quantitatively as

\begin{eqnarray}
  L^{FL} \approx 1.19,~\mbox{and}~L^{CE} \approx 1.23,
\end{eqnarray}

\noindent where the two values are very close. Since the loss scale can reflect the gradient magnitude, we believe that the similar gradient magnitude is the reason for \textit{RetinaNet-CE} with $s=10^{-1},\pi=10^{-5}$ achieves AP comparable to that of \textit{RetinaNet-FL}. Moreover, this setting also facilitates multi-task learning. See Fig.~\ref{figure3a}, \textit{RetinaNet-FL} achieves a quite good balance between the localization loss $L^{LOC}$ and $L^{FL}$, whereas Fig.~\ref{figure3b} shows that only \textit{RetinaNet-CE} with $\pi=10^{-5}, w=10^{-1}$ can obtain this balance. Other settings, however, will lead the training to be dominated by the localization task, which is harmful to training an accurate object detector. In conclusion, our investigation reveals that the classification gradient magnitude is the central cause of the accuracy gap, and we can adjust it by initialization and loss scaling. 

\section{Method}\label{section4}

Our investigation reveals that by tuning $s$ and $\pi$, the detector without heuristic sampling methods can achieve a similar detection accuracy to that with heuristic sampling methods. However, tuning them is always laborious. In this section, we propose a novel Sampling-Free mechanism, which addresses the \textit{fg-bg} imbalance by adaptively setting $s$ and $\pi$, thus adaptively controlling the classification gradient magnitude.

\textit{1) Discarding Heuristic Sampling Methods:} 
As sampling methods are always a default part in training deep object detectors, the first step of our Sampling-Free mechanism is discarding heuristic sampling methods during training. For one-stage anchor-based object detectors~\cite{ron,focal_loss,yolov2,yolov3,refinedet,rfbnet,fsaf,freeanchor,mal}, soft sampling methods (\eg Focal Loss~\cite{focal_loss}, GHM~\cite{ghm}, PISA~\cite{pisa}) is widely used for re-weighting training samples in the classification task. In our method, we use the standard CE loss to train the classification task, which treats all training samples equally.

For two-stage anchor-based approaches~\cite{faster_rcnn,rfcn,fpn,mask_rcnn,relod,cascade_rcnn,iounet,grid_rcnn,libra_rcnn,htc,garpn,tridentnet}, hard sampling methods (\eg biased sampling~\cite{faster_rcnn}, OHEM~\cite{ohem}) are widely used for re-sampling training samples. In our method, we train all training samples in RPN and RoI-subnet. For example, a common implementation of biased sampling~\cite{faster_rcnn} in training Faster R-CNN is: (1) RPN randomly selects 256 anchors with a biased 1:1 \textit{fg-to-bg} ratio, (2) RoI-subnet randomly selects 512 proposals with a biased 1:3 \textit{fg-to-bg} ratio. In our method, we train all examples in both RPN and RoI-subnet. That is to say, we train RPN with all foreground/background anchors ($\sim$$10^5$ per-image) and train RoI-subnet with all foreground/background proposals ($\sim$$10^3$ per-image). We use CE loss as the classification loss both in RPN and RoI-subnet.

For anchor-free object detectors~\cite{cornernet,extremenet,grid_rcnn,centernet_triplets,reppoints,dense_reppoints,fcos,garpn,foveabox,sapd}, they regard ``points'' as the training samples rather than ``anchors'' in anchor-based object detectors, and most of them use soft sampling methods to address the \textit{fg-bg} imbalance. We follow the principle of Sampling-Free in anchor-based object detectors that equally use all training samples during training, \ie we use the standard CE loss in the classification task.

\textit{2) Optimal Bias Initialization:} 
Sec.~\ref{section3.3} has shown that adjusting $\pi$ can help to avoid network diverging. However, it is difficult to determine $\pi$. We propose optimal bias initialization to compute $\pi$ from data statistics rather than tuning it. Our idea is to initialize the bias of the last convolutional layer to obtain a minimal classification loss value. The derivative of $L^{CE}$ is

\begin{eqnarray}
    \frac{\partial L^{CE}}{\partial \pi} = -\frac{1}{\pi} + (\frac{N}{N^f} - 1)\frac{1}{1 - \pi}.
 \end{eqnarray}

 \noindent When $\pi = \frac{N^f}{N}$, $\frac{\partial L^{CE}}{\partial \pi}=0$, and $L^{CE}$ will attain the minimal value. As the predicted score is predicted with the sigmoid activation, we can obtain the optimal initial bias as

 \begin{eqnarray}
    b = -\log\frac{1-\pi}{\pi} = -\log(\frac{N}{N^f}- 1).
 \end{eqnarray}

 \noindent Here $\frac{N}{N^f}$ can be computed by pre-defined anchors, thus the computation is efficient as it does not require network forwarding. In our observation, $\frac{N}{N^f}\approx10^{5}$, which corresponds to $\pi=10^{-5}$ that performs best in our experiments. It is worth noting that the accuracy of the model is robust to our initialization strategy, as the model can ``utilize'' the imbalanced distribution to obtain a lower loss. We initialize the model to ensure the stability of the classification loss.

\textit{3) Guided Loss Scaling:}
Usually, the overall loss function to train a deep object detector is composed of a localization loss term $L^{LOC}$ and a classification loss term $L^{CLS}$. Let $L^t$ denote the overall loss in the $t$-th training step. When we use CE loss as the classification loss, we have 

\begin{equation}
  L^t = (L^{LOC})^t + (L^{CLS})^t = (L^{LOC})^t + s^t(L^{CE})^t,
\end{equation}

\noindent where $s^t$ is used to scale the $(L^{CE})^t$ as the CE loss scale is unreasonable under the \textit{fg-bg} imbalance. As mentioned in \ref{section3.3}, it is essential to control the classification loss scale to be close to the localization loss scale. A straightforward way is to adjust $s^t$. However, it results in a new hyperparameter. Our key idea is to adjust $s^t$ dynamically during training. That says, instead of using a constant $s^t$, we define a guided term

\begin{equation}
g^t =\frac{(L^{LOC})^t}{(L^{CE})^t}, 
\end{equation}

\noindent and let $s^t = g^t$, which suggests using the localization loss scale of the current mini-batch as the target of the rescaled CE loss scale. Thus, this technique is termed ``guided loss scaling''. It is worth noting that $g^t$ is only used for scaling the classification loss, \ie its gradient is ignored in the backpropagation. Therefore, the overall gradient is 

\begin{equation}
  \frac{\partial L^t}{\partial \Theta^t} = \frac{\partial (L^{LOC})^t}{\partial \Theta^t} + g^t\frac{\partial (L^{CE})^t}{\partial \Theta^t},
\end{equation}

\noindent which ignores the gradient calculation of $s^t$. 

Our guided loss scaling can be interpreted threefold. First, according to Sec.~\ref{section3.3} (especially Fig.~\ref{figure3}), it appears a good choice to let the localization loss scale and the classification loss scale be similar, where the classification loss is either Focal Loss or CE loss. Second, it is convenient to use the localization loss as guidance, because the localization loss is already there for object detection. Third, the classification loss without sampling methods (\ie CE loss) is greatly influenced by the \textit{fg-bg} imbalance, but localization loss is little influenced as it is computed merely for foreground anchors. Thus, the localization loss is helpful to control the unreasonable classification loss due to the \textit{fg-bg} imbalance.

\begin{table*}[t]
  \scriptsize
  \centering
  \caption{Ablation studies of our Sampling-Free mechanism on COCO \texttt{minival}. ``FL $\rightarrow$ CE'' denotes ``Focal Loss $\rightarrow$ CE loss'', and ``Init'' denotes ``Initialization''. From the following three subtables, better AP can be achieved with Sampling-Free, but the hybrid of the guided loss scaling and heuristic sampling methods cannot improve the detection accuracy.}
  \subtable[Ablation studies of Sampling-Free in RetinaNet~\cite{focal_loss}]{
      \begin{tabular}{|c|c|c|c|c|c|c|c|c|c|}
      \hline 
      Components & \multicolumn{6}{c|}{RetinaNet (ResNet-50-FPN, 1$\times$)} \\
      \hline 
      FL $\rightarrow$ CE & \xmark & \cmark & \cmark & \cmark & \xmark & \cmark \\
      \hline
      Optimal Bias Init & \xmark & \xmark & \cmark & \xmark & \cmark & \cmark \\
      \hline
      Guided Loss Scaling & \xmark & \xmark & \xmark & \cmark & \cmark & \cmark \\
      \hline\hline 
      AP        & 36.4 & n/a & n/a & n/a & 36.5 (+0.1) & \textbf{37.0 (+0.6)} \\ 
      AP$_{50}$ & 55.0 & n/a & n/a & n/a & 55.5 (+0.5) & \textbf{56.5 (+1.5)} \\
      AP$_{75}$ & 39.0 & n/a & n/a & n/a & 38.8 (-0.2) & \textbf{39.2 (+0.2)} \\
      AP$_{S}$  & 19.9 & n/a & n/a & n/a & 20.1 (+0.2) & \textbf{20.3 (+0.4)} \\
      AP$_{M}$  & 40.3 & n/a & n/a & n/a & 40.1 (-0.2) & \textbf{40.5 (+0.2)} \\
      AP$_{L}$  & 48.9 & n/a & n/a & n/a & 48.1 (-0.8) & \textbf{49.5 (+0.6)} \\
      \hline
      \end{tabular}\label{table3a}
  }
  \subtable[Ablation studies of Sampling-Free in FCOS~\cite{fcos}]{
      \begin{tabular}{|c|c|c|c|c|c|c|c|c|c|}
      \hline 
      Components & \multicolumn{6}{c|}{FCOS (ResNet-50-FPN, 1$\times$)} \\
      \hline 
      FL $\rightarrow$ CE & \xmark & \cmark & \cmark & \cmark & \xmark & \cmark \\
      \hline
      Optimal Bias Init & \xmark & \xmark & \cmark & \xmark & \cmark & \cmark \\
      \hline
      Guided Loss Scaling & \xmark & \xmark & \xmark & \cmark & \cmark & \cmark \\
      \hline\hline 
      AP        & 37.1 & n/a & n/a & n/a & 37.1 (+0.0) & \textbf{37.6 (+0.5)} \\ 
      AP$_{50}$ & 56.0 & n/a & n/a & n/a & 56.2 (+0.2) & \textbf{57.4 (+1.4)} \\
      AP$_{75}$ & 39.8 & n/a & n/a & n/a & 39.7 (-0.1) & \textbf{40.3 (+0.5)} \\
      AP$_{S}$  & 21.3 & n/a & n/a & n/a & 21.0 (-0.3) & \textbf{21.9 (+0.6)} \\
      AP$_{M}$  & 41.0 & n/a & n/a & n/a & 41.3 (+0.3) & \textbf{41.2 (+0.2)} \\
      AP$_{L}$  & 47.8 & n/a & n/a & n/a & 47.9 (+0.1) & \textbf{48.5 (+0.7)} \\
      \hline
      \end{tabular}\label{table3b}
  }
  \subtable[Ablation studies of Sampling-Free in Faster R-CNN~\cite{faster_rcnn}]{
    \begin{tabular}{|c|cc|cc|cc|cc|cc|cc|cc|}
      \hline 
      \multirow{2}*{Sampling-Free Mechanism} & \multicolumn{14}{|c|}{Faster R-CNN (ResNet-50-FPN~\cite{resnet,fpn}, 1$\times$)} \\
      \cline{2-15}
                            & RPN & RoI & RPN & RoI & RPN & RoI & RPN & RoI & RPN & RoI & RPN & RoI & RPN & RoI \\
      \hline 
      Biased Sampling $\rightarrow$ Non-sampling & \xmark & \xmark & \xmark & \xmark & \xmark & \xmark & \xmark & \xmark & \cmark & \xmark & \xmark & \cmark & \cmark & \cmark \\
      Optimal Bias Initialization + Guided Loss Scaling                               & \xmark & \xmark & \cmark & \xmark & \xmark & \cmark & \cmark & \cmark & \cmark & \xmark & \xmark & \cmark & \cmark & \cmark \\
      \hline\hline 
      AP        & \multicolumn{2}{|c|}{36.8} & \multicolumn{2}{|c|}{36.5 (-0.3)}& \multicolumn{2}{|c|}{36.4 (-0.4)}&\multicolumn{2}{|c|}{36.8 (+0.0)}& \multicolumn{2}{|c|}{37.5 (+0.7)} &\multicolumn{2}{|c|}{38.1 (+1.3)} & \multicolumn{2}{|c|}{\textbf{38.4 (+1.6)}}\\
      AP$_{50}$ & \multicolumn{2}{|c|}{58.4} & \multicolumn{2}{|c|}{58.2 (-0.2)}& \multicolumn{2}{|c|}{57.9 (-0.5)}&\multicolumn{2}{|c|}{58.7 (+0.3)}& \multicolumn{2}{|c|}{59.0 (+0.6)} &\multicolumn{2}{|c|}{59.6 (+1.2)} & \multicolumn{2}{|c|}{\textbf{59.9 (+1.5)}}\\
      AP$_{75}$ & \multicolumn{2}{|c|}{40.0} & \multicolumn{2}{|c|}{39.6 (-0.4)}& \multicolumn{2}{|c|}{39.4 (-0.6)}&\multicolumn{2}{|c|}{40.0 (+0.0)}& \multicolumn{2}{|c|}{40.4 (+0.4)} &\multicolumn{2}{|c|}{41.6 (+1.6)} & \multicolumn{2}{|c|}{\textbf{41.7 (+1.7)}}\\
      AP$_{S}$  & \multicolumn{2}{|c|}{20.7} & \multicolumn{2}{|c|}{21.2 (+0.5)}& \multicolumn{2}{|c|}{21.1 (+0.4)}&\multicolumn{2}{|c|}{21.1 (+0.4)}& \multicolumn{2}{|c|}{21.5 (+0.8)} &\multicolumn{2}{|c|}{22.2 (+1.5)} & \multicolumn{2}{|c|}{\textbf{22.3 (+1.6)}}\\
      AP$_{M}$  & \multicolumn{2}{|c|}{39.7} & \multicolumn{2}{|c|}{39.4 (-0.3)}& \multicolumn{2}{|c|}{39.1 (-0.6)}&\multicolumn{2}{|c|}{40.0 (+0.3)}& \multicolumn{2}{|c|}{40.7 (+1.0)} &\multicolumn{2}{|c|}{41.2 (+1.5)} & \multicolumn{2}{|c|}{\textbf{41.6 (+1.9)}}\\
      AP$_{L}$  & \multicolumn{2}{|c|}{47.9} & \multicolumn{2}{|c|}{47.6 (-0.3)}& \multicolumn{2}{|c|}{47.5 (-0.4)}&\multicolumn{2}{|c|}{47.8 (-0.1)}& \multicolumn{2}{|c|}{48.8 (+0.9)} &\multicolumn{2}{|c|}{50.0 (+2.1)} & \multicolumn{2}{|c|}{\textbf{50.9 (+3.0)}}\\
      \hline
      \end{tabular}\label{table3c}
  }
  \label{table3}
\end{table*}

However, in our experiments, we find that the detector may not achieve the best detection accuracy when the classification loss is simply equal to the localization loss. Fortunately, the well-known uncertainty weighting~\cite{uncertainty} proposes a simple method to weigh two losses from the perspective of Bayesian uncertainty. When we apply the method to our case, the overall loss would be

\begin{eqnarray}
  L^t = \frac{1}{(\sigma_1^t)^2}(L^{LOC})^t + \frac{1}{(\sigma_2^t)^2}(L^{CE})^t + 2\log(\sigma_1^t\sigma_2^t),
\end{eqnarray}

\noindent where $\sigma_1^t$ and $\sigma_2^t$ are learnable parameters, and they are initialized as $\sigma_1^0 = \sigma_2^0 = 1$. $2\log(\sigma_1^t\sigma_2^t)$ is the normalization term to avoid the degradation of $\frac{1}{(\sigma_1^t)^2} \rightarrow 0$ and $\frac{1}{(\sigma_2^t)^2} \rightarrow 0$. But if we train the detector in this way, the training would be quickly failed, as the classification loss would be much larger ($\sim$\text10$\times$) than the localization loss at the start of the training (see Figure~\ref{figure3}). Hence, the guided term $g^t$ is necessary, and the overall loss should be

\begin{eqnarray}
  L^t = \frac{1}{(\sigma_1^t)^2}(L^{LOC})^t + \frac{g^t}{(\sigma_2^t)^2}(L^{CE})^t + 2\log(\sigma_1^t\sigma_2^t).
\end{eqnarray}

\noindent To keep the consistency with the original training loss, we can identify that the localization loss does not require weighting ($\sigma_1^t=1$). We can also use $\delta^t$ to denote $\frac{1}{(\sigma_2^t)^2}$, then the overall loss would be very simple, \ie  

\begin{eqnarray}
  L^t = (L^{LOC})^t + g^t\delta^t(L^{CE})^t - \log\delta^t.
\end{eqnarray}

We notice that there have been several works~\cite{uncertainty,gradnorm,mtl_as_moo} for adaptive multi-task loss scaling. Our guided loss scaling is different from them in three points: 1) it is aimed at controlling the classification loss under the \textit{fg-bg} imbalance, which belongs to the single-task loss weighting rather than the multi-task loss weighting; 2) it is specifically designed for deep object detectors as it requires the localization loss to guide the classification loss; 3) it converts the class imbalance problem to the loss scaling problem, which seems not reported before in the literature, to our best knowledge.

\section{Experiments}
In this section, we will perform extensive experiments to validate our Sampling-Free mechanism. Before that, we first describe the experimental details about datasets and baselines. Then, we perform ablation studies on anchor-based and anchor-free object detectors. Finally, we compare our method with existing heuristic sampling methods, and present experimental results on public datasets.

\subsection{Implementation Details}

\textit{1) Datasets:} 
We use the well-known COCO~\cite{coco} and PASCAL VOC~\cite{pascal_voc} datasets to validate our method. For COCO dataset, following common practices~\cite{focal_loss,faster_rcnn}, we train models on the \texttt{train2017} split and perform ablation studies on \texttt{minival} split, and report detection accuracy on \texttt{test-dev} split, where COCO-style average precision (AP) is used as the evaluation metrics. For PASCAL VOC dataset, we also follow common practices~\cite{faster_rcnn,refinedet} that training models on a union set of PASCAL VOC 2007 and 2012 set (\texttt{07+12} split), and evaluated on PASCAL VOC 2007 test set (\texttt{07test} split), where VOC-style mean average precision (mAP) is used as the evaluation metrics.

\textit{2) Baselines:} 
We use three object detectors --- RetinaNet~\cite{focal_loss} (one-stage anchor-based), Faster R-CNN~\cite{faster_rcnn} (two-stage anchor-based), FCOS~\cite{fcos} (anchor-free) that implemented on \texttt{maskrcnn-benchmark}~\cite{maskrcnn_benchmark} to perform experiments, where we follow the public standard training configurations to implement them, which means that we have not made any changes for the hyperparameters of them. Besides, we also use Mask R-CNN~\cite{mask_rcnn} to validate Sampling-Free in the instance segmentation task.

\subsection{Ablation Studies}

\begin{figure*}[t]
  \centering
  \includegraphics[width=\linewidth]{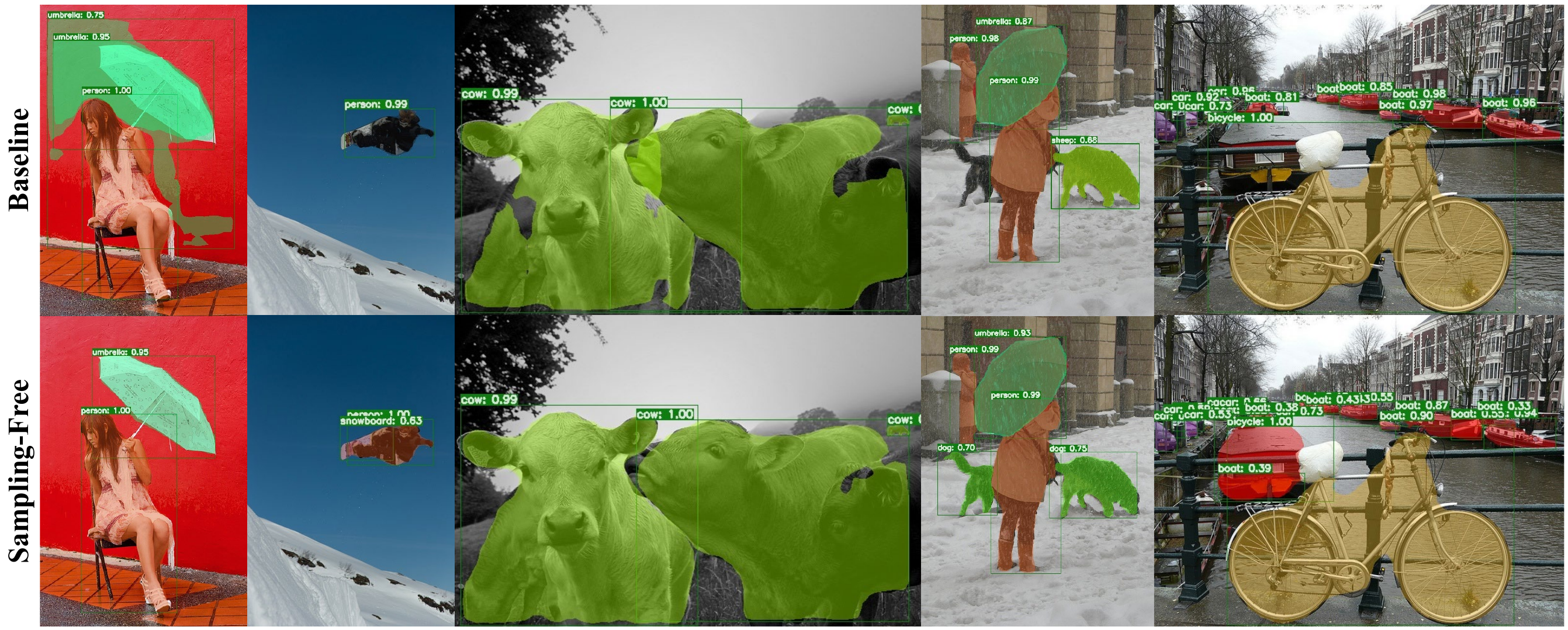}
  \caption{Mask R-CNN~\cite{mask_rcnn} (37.8 box AP, 34.2 mask AP on COCO \texttt{minival}) vs. Mask R-CNN with Sampling-Free (39.0 box AP, 34.9 mask AP on COCO \texttt{minival}) in ResNet-50-FPN backbone. The latter exhibits better detection and segmentation results. }
  \label{figure4}
\end{figure*}

\textit{1) Detection Accuracy:} 
As shown in Table~\ref{table3}, we perform experiments of Sampling-Free on RetinaNet, FCOS, and Faster R-CNN, respectively. See Table~\ref{table3a} and Table~\ref{table3b}, the experimental phenomena of Sampling-Free on RetinaNet and FCOS are similar. Only when we discard heuristic sampling methods (Focal Loss $\rightarrow$ CE Loss) and use optimal bias initialization with guided loss scaling at the same time, we can obtain meaningful detection accuracy improvements. Specifically, Sampling-Free achieves 0.6 AP and 0.5 AP higher than Focal Loss in RetinaNet and FCOS, respectively. This is an impressive improvement as the well-know GHM-C~\cite{ghm} only outperforms Focal Loss 0.2 AP. More importantly, this is the first time that CE Loss has achieved better performance than the soft sampling method in object detection.

For Faster R-CNN (See Table~\ref{table3c}), when we discard heuristic sampling methods without optimal bias initialization and guided loss scaling, the training of the detector will be failed. When we only use the optimal bias initialization and guided loss scaling without discarding heuristic sampling methods, the detector also cannot obtain improvement on detection accuracy. Only when we discard heuristic sampling methods, and use optimal bias initialization with guided loss scaling at the same time, we will observe obvious gains in AP. Sampling-Free improves 0.7 AP and 1.3 AP in RPN and RoI-subnet, respectively. When we use Sampling-Free in both RPN and RoI-subnet, an impressive 1.6 AP improvement can be obtained, with the gains from all AP metrics. 

We notice that Sampling-Free exhibits more improvements to Faster R-CNN than RetinaNet and FCOS, which may be due to biased sampling wasting most background examples, but Sampling-Free allows all foreground and background examples to be trained synchronously.

\begin{table}[t]
  \small
  \centering
  \caption{Training speed and memory cost of Sampling-Free on Faster R-CNN, which is evaluated on a single \texttt{Nvidia-Titan-Xp} GPU with batch size 1.}
  \begin{tabular}[t]{|c|c|c|c|}
      \hline
      Detector & Method & Speed & Memory \\
      \hline
      Faster R-CNN & Biased Sampling & \textbf{172ms} & 1714MB \\
      \cline{2-4}
      (ResNet-50-FPN, 1$\times$) & Sampling-Free & 184ms & \textbf{1669MB} \\
      \hline
  \end{tabular}
  \label{table4}
\end{table}

\textit{2) Training Speed and Memory Cost:} 
As both Focal Loss and Sampling-Free use all samples to train, the training speed and the memory cost of \textit{RetinaNet-FL} and \textit{RetinaNet-CE} are very similar. However, in Faster R-CNN, Sampling-Free allows more background samples to be trained. As shown in Table~\ref{table4}, we measure the performance of Faster R-CNN with Sampling-Free on a single \texttt{Nvidia-Titan-Xp} GPU with batch size 1. Interestingly, although the training speed becomes slower for Faster R-CNN with Sampling-Free (as more background proposals are trained in RoI-subnet), its memory cost is reduced, which is owing to the operation of biased sampling also requires considerable memory costs.

\begin{table}[t] 
  \footnotesize 
  \centering
  \caption{Results of Sampling-Free on COCO \texttt{test-dev}.  }\label{table5}
  \begin{tabular}[t]{|c|c|ccc|}
      \hline
      Detector & Method & AP & AP$_{50}$ & AP$_{75}$   \\ 
      \hline
      Faster R-CNN & Biased  Sampling & 39.3 & 61.4 & 42.7 \\
      (ResNet-101-FPN, 1$\times$) & Sampling-Free & \textbf{40.7} & 62.4 & 44.3 \\
      \cline{1-2}\cline{3-5}
      RetinaNet  & Focal Loss & 38.8 & 58.4 & 41.7 \\
      (ResNet-101-FPN, 1$\times$) & Sampling-Free & \textbf{39.4} & 58.2 & 42.9 \\
      \hline
      Faster R-CNN & Biased  Sampling & 39.6 & 61.3 & 43.0 \\
      (ResNet-101-FPN, 2$\times$) & Sampling-Free & \textbf{41.0} & 62.6 & 44.6 \\
      \cline{1-2}\cline{3-5}
      RetinaNet & Focal Loss & 38.9 & 58.4 & 41.7 \\
      (ResNet-101-FPN, 2$\times$) & Sampling-Free & \textbf{39.5} & 58.3 & 43.2 \\
      \hline
  \end{tabular}
\end{table}

\begin{table}[t]
  \centering
  \small
  \caption{Results of Sampling-Free on PASCAL VOC \texttt{07test} split.}\label{table6}
  \begin{tabular}{|c|c|c|}
    \hline
    Detector & Method & mAP \\
    \hline
    RetinaNet  & Focal Loss & 79.3 \\
    \cline{2-3}
    (ResNet-50-FPN, 0.2$\times$)   & Sampling-Free & \textbf{80.1} \\
    \hline
    Faster R-CNN & Biased  Sampling & 80.9 \\
    \cline{2-3}
    (ResNet-50-FPN, 0.2$\times$)  & Sampling-Free & \textbf{81.5} \\
    \hline
  \end{tabular}
\end{table}

\begin{table*}[t]
  \footnotesize
  \centering 
  \caption{This table illustrates the comparison between heuristic sampling methods and our Sampling-Free mechanism. $\Delta$AP and $\Delta$hyperparameters denote the change in detection accuracy and hyperparameters relative to the baseline method (Focal Loss in RetinaNet, Biased  Sampling in Faster R-CNN). Sampling-Free achieves the best $\Delta$AP  without any hyperparameter introduced.} \label{table7}
  \begin{tabular}{|c|c|c|c|c|}
  \hline
  Solutions & Abbreviation & $\Delta$AP in RetinaNet (R-50-FPN, 1$\times$) & $\Delta$AP in Faster R-CNN (R-50-FPN, 1$\times$) & $\Delta$hyperparameters \\
  \hline 
  \multirow{2}*{Hard Sampling} & OHEM~\cite{ohem} & n/a & 36.4 $\rightarrow$ 36.6 (+0.2 AP) & 2 $\rightarrow$ 2 \\
  \cline{2-5}
  ~ & IoU-balanced sampling~\cite{libra_rcnn} & n/a & 36.4 $\rightarrow$ 36.8 (+0.4 AP) & 2 $\rightarrow$ 3 \\
  \hline
  \multirow{2}*{Soft Sampling} & GHM-C~\cite{ghm} & 35.6 $\rightarrow$ 35.8 (+0.2 AP) & n/a & 2 $\rightarrow$ 1 \\
  \cline{2-5}
  ~ & ISR~\cite{pisa} & n/a & 36.4 $\rightarrow$ 37.9 (+1.5 AP) & 2 $\rightarrow$ 4 \\ 
  \hline
  Non-Sampling & Sampling-Free & \textbf{36.4 $\rightarrow$ 37.0 (+0.6 AP)} & \textbf{36.8 $\rightarrow$ 38.4 (+1.6 AP)} & \textbf{2 $\rightarrow$ 0} \\
  \hline
  \end{tabular} 
\end{table*}

\begin{table}[t]
  \centering
  \caption{Results of adaptive label assignment strategies with our Sampling-Free mechanism.}\label{table8}
  \subtable[AP of ATSS (ResNeXt-64x4d-101-DCN, 2$\times$) on COCO \texttt{test-dev}]{
    \begin{tabular}[t]{|c|c|ccccc|}
        \hline
        Method & AP & AP$_{50}$ & AP$_{75}$ & AP$_{S}$ & AP$_{M}$ & AP$_{L}$  \\ 
        \hline
        ATSS & 47.7 & 66.5 & 51.9 & 29.7 & 50.8 & 59.4 \\
        \hline
        w. Sampling-Free & \textbf{48.2} & \textbf{66.9} & \textbf{52.4} & \textbf{30.3} & \textbf{51.3} & \textbf{59.9} \\
        \hline
    \end{tabular}
    }
    \subtable[AP of PAA (ResNeXt-64x4d-101-DCN, 2$\times$) on COCO \texttt{test-dev}]{
        \begin{tabular}[t]{|c|c|ccccc|}
            \hline
            Method & AP & AP$_{50}$ & AP$_{75}$ & AP$_{S}$ & AP$_{M}$ & AP$_{L}$  \\ 
            \hline
            PAA & 49.0 & 67.8 & 53.3 & 30.2 & 52.8 & 62.2 \\
            \hline
            w. Sampling-Free & \textbf{49.6} & \textbf{68.3} & \textbf{53.8} & \textbf{30.6} & \textbf{53.7} & \textbf{63.1} \\
            \hline
        \end{tabular}
    }
\end{table}

\subsection{Experimental Results}
\textit{1) Results on COCO and PASCAL VOC:} 
For COCO dataset, we have demonstrated that the effectiveness of Sampling-Free on ResNet-50-FPN backbone and 1$\times$ learning schedule. We further verify our methods on the larger backbone and the longer learning schedule. As shown in Table~\ref{table5}, for ResNet-101-FPN backbone, Sampling-Free still shows impressive detection accuracy improvements, which can improve Faster R-CNN and RetinaNet about 1.5 AP and 0.5 AP, respectively. Even with the $2\times$ learning schedule, we observe a steady increase in AP as well. 

For PASCAL VOC dataset, as shown in Table~\ref{table6}, Sampling-Free improves 0.8 mAP and 0.6 mAP for RetinaNet and Faster R-CNN, respectively. These results illustrate the robustness of our Sampling-Free mechanism.

\begin{table}[t]
  \footnotesize
  \centering
  \caption{Results of Sampling-Free in the instance segmentation task.}\label{table9}
  \subtable[Box AP on COCO \texttt{minival} (ResNet-50-FPN, 1$\times$) ]{
    \begin{tabular}[t]{|c|c|ccccc|}
        \hline
        Method & AP & AP$_{50}$ & AP$_{75}$ & AP$_{S}$ & AP$_{M}$ & AP$_{L}$  \\ 
        \hline
        Mask R-CNN & 37.8 & 59.3 & 41.1 & 21.5 & 41.1 & 49.9 \\
        \hline
        w. Sampling-Free & \textbf{39.0} & \textbf{60.3} & \textbf{42.5} & \textbf{22.5} & \textbf{41.9} & \textbf{51.2} \\
        \hline
    \end{tabular}
    }
    \subtable[Mask AP on COCO \texttt{minival} (ResNet-50-FPN, 1$\times$)]{
        \begin{tabular}[t]{|c|c|ccccc|}
            \hline
            Method & AP & AP$_{50}$ & AP$_{75}$ & AP$_{S}$ & AP$_{M}$ & AP$_{L}$  \\ 
            \hline
            Mask R-CNN & 34.2 & 55.9 & 36.3 & 15.6 & 36.8 & 50.6 \\
            \hline
            w. Sampling-Free & \textbf{34.9} & \textbf{56.8} & \textbf{37.1} & \textbf{16.2} & \textbf{37.3} & \textbf{51.2} \\
            \hline
        \end{tabular}
    }
\end{table}

\begin{figure*}[t]
  \centering
  \subfigure[RetinaNet with Sampling-Free exhibits better performance than RetinaNet with Focal Loss~\cite{focal_loss} (baseline)]{
      \includegraphics[width=0.975\linewidth]{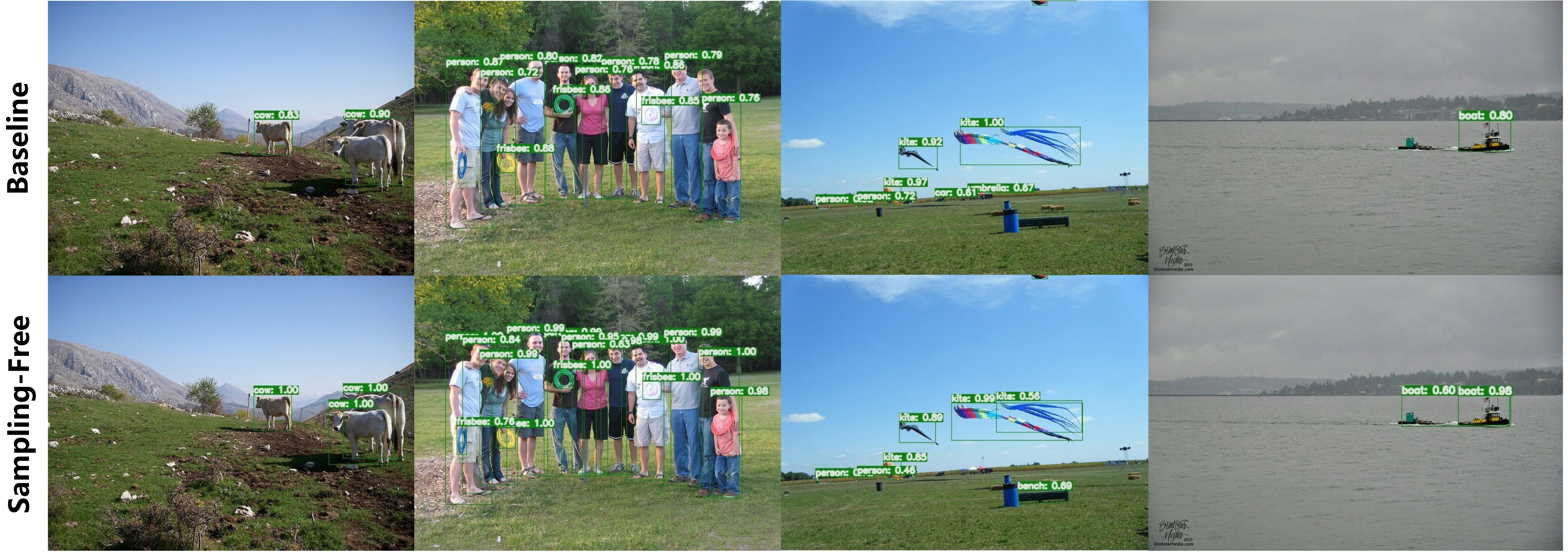}
  }
  \subfigure[Faster R-CNN with Sampling-Free exhibits better performance than Faster R-CNN with biased sampling~\cite{faster_rcnn} (baseline)]{
      \includegraphics[width=0.975\linewidth]{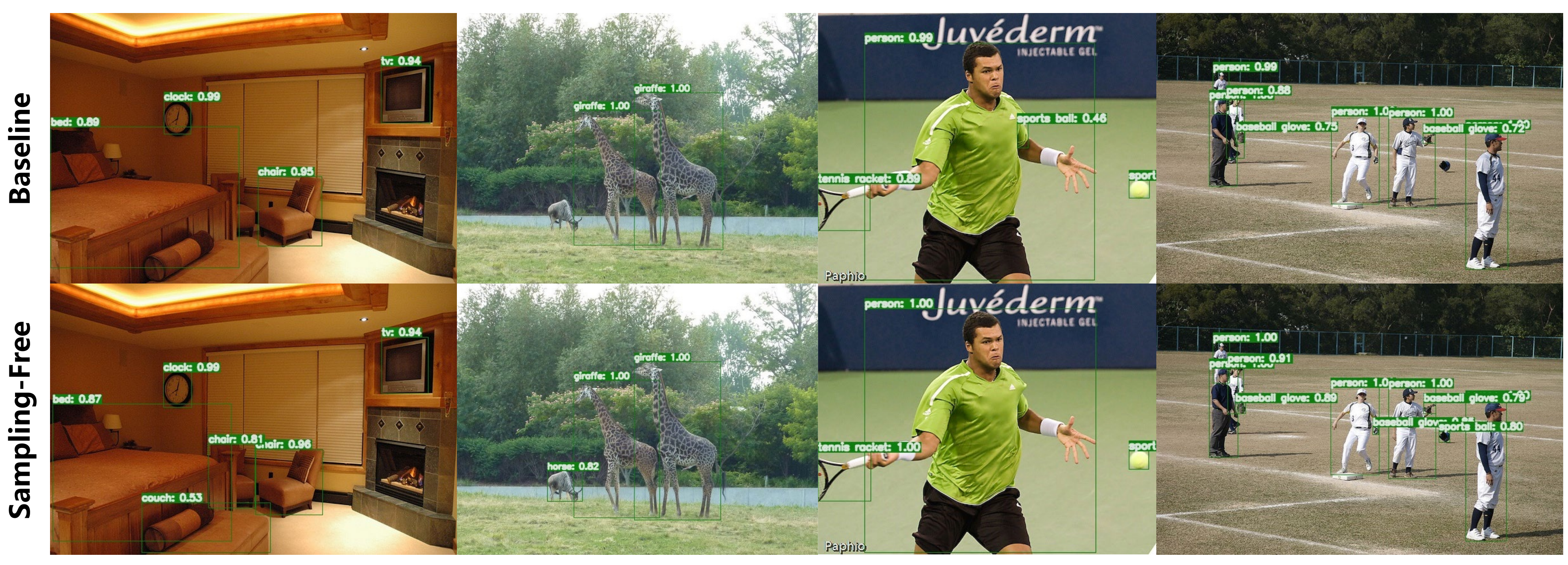}
  }
  \caption{Visualized detection results of different object detectors with and without our Sampling-Free mechanism, which show that the detector with Sampling-Free mechanism performs better. The backbone of these detectors is ResNet-50-FPN~\cite{resnet,fpn}.}\label{figure5}
\end{figure*}

\textit{2) Comparison:} 
We compare Sampling-Free with existing heuristic sampling methods to illustrate our advantages. As the implementations of Sampling-Free and other methods may be on different platforms (\eg \texttt{maskrcnn-benchmark} vs. \texttt{mmdetection}), we mainly compare their changes in performance relative to the baseline method (Focal Loss in RetinaNet, Biased  Sampling in Faster R-CNN). As presented in Table~\ref{table7}, compared with heuristic sampling methods, Sampling-Free has the following three advantages:

$\bullet$ Sampling-Free achieves the best relative detection accuracy improvement, where 0.6 AP and 1.6 AP improvement are obtained in RetinaNet and Faster R-CNN, respectively.

$\bullet$ Sampling-Free has no hyperparameters to search --- In contrast, other heuristic sampling methods have at least one hyperparameter. The ISR~\cite{pisa}, which is closest to us in detection accuracy, introduces 4 hyperparameters.

$\bullet$ Sampling-Free has better versatility --- it is applicable to both one-stage and two-stage deep object detectors.

\textit{3) Results on Adaptive Label Assignment Strategies:} To date, the state-of-the-art detection accuracy is achieved by adaptive label assignment methods~\cite{atss,paa}, where the definition of foreground/background training sample is very different from that in deep object detectors. To validate the effectiveness of Sampling-Free in them, we replace Focal Loss with Sampling-Free in ATSS~\cite{atss} and PAA~\cite{paa}. As presented in Table~\ref{table8}, we successfully verify that the state-of-the-art models of ATSS and PAA can be further improved with Sampling-Free.

\textit{4) Instance Segmentation:} Although Sampling-Free can help detectors achieve better detection accuracy, it is unknown whether detection results produced with Sampling-Free mechanism can facilitate other tasks in practice. Therefore, we introduce Sampling-Free into Mask R-CNN to observe whether it can achieve higher accuracy in instance segmentation. Note that in Mask R-CNN, the heuristic sampling method is not used in its segmentation branch, thus we only apply our Sampling-Free mechanism on the classification branch. As shown in Table~\ref{table9}, Mask R-CNN with Sampling-Free achieves 1.2 box AP and 0.7 mask AP gains. We visualize the detection and segmentation results in Fig.~\ref{figure5}, which suggests that the detection results produced with Sampling-Free can also improve the instance segmentation task.

\textit{5) Visualization:} 
More visualization results are in Fig.~\ref{figure5}.

\section{Conclusion}
In this paper, we explored whether heuristic sampling methods are necessary for training accurate deep object detectors under the \textit{fg-bg} imbalance. Our investigation revealed that the key to training without heuristic sampling methods under the \textit{fg-bg} imbalance is to control the classification gradient magnitude. Inspired by this, we proposed a novel Sampling-Free mechanism to control the classification gradient magnitude from initialization and loss scaling, without new hyperparameters introduced. Extensive experiments demonstrated the effectiveness of Sampling-Free in various object detectors, which also yields considerable gains in the instance segmentation task and the state-of-the-art label assignment strategies. Our Sampling-Free mechanism provides a new perspective to address the \textit{fg-bg} imbalance.

Although Sampling-Free can support training with cross-entropy loss, it is not designed for the detection metrics, which may limit further performance increases. Specifically, the average precision metric expects samples to have an IoU-related confidence score. There has been some work on this~\cite{varifocalnet,gfl}, but they still use a variant of Focal Loss to train the classification task. Thus, a future study direction is to use only cross-entropy loss to model this unified confidence score. On a larger scale, one can try to use cross-entropy loss with metric-specific design to achieve better performance.

\bibliographystyle{IEEEtran}
\bibliography{ai.bib}

\newpage

\begin{IEEEbiography}[{\includegraphics[width=1in,height=1.25in,clip,keepaspectratio]{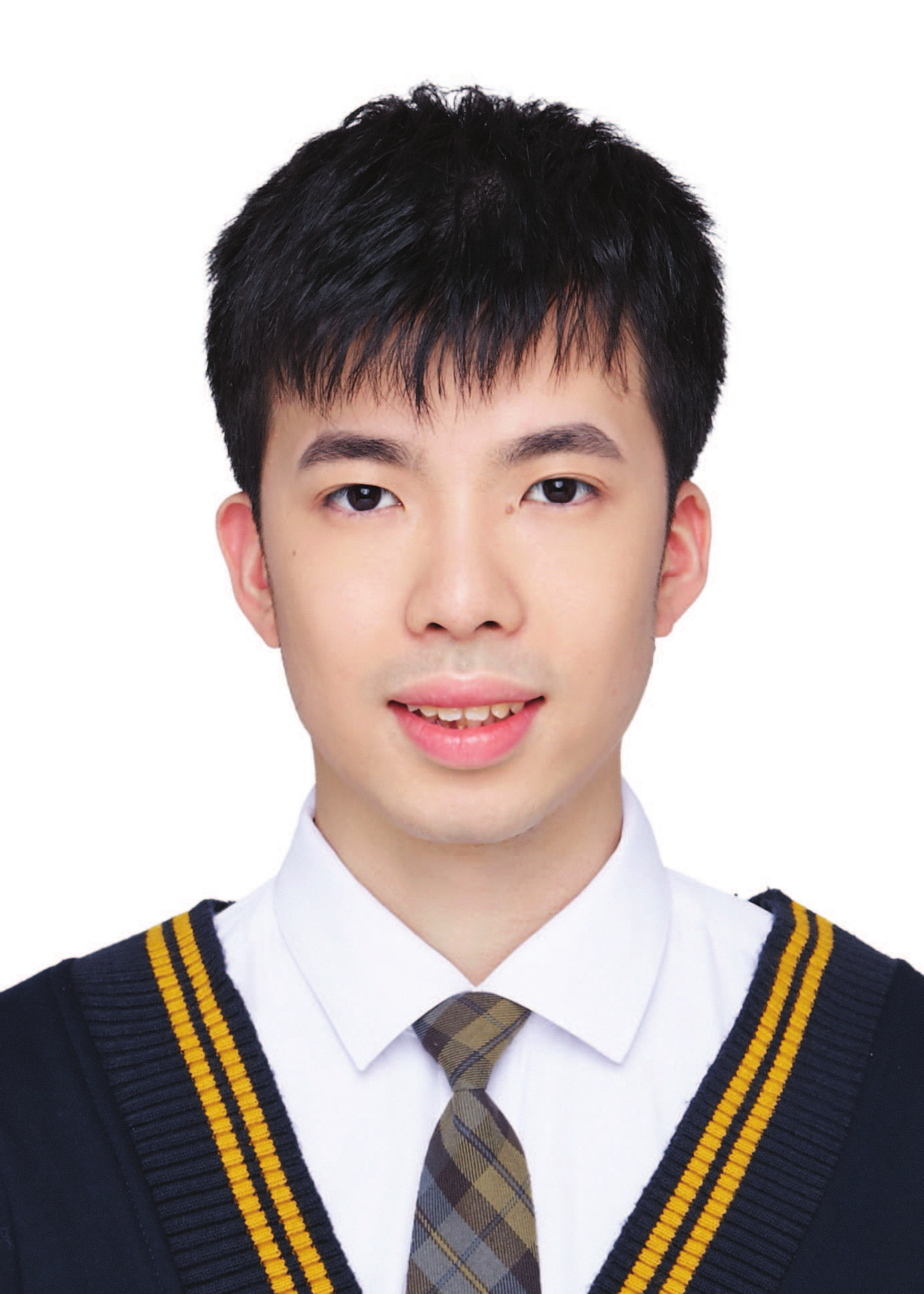}}]{Joya Chen} 
  (M'20) received his bachelor's degree in School of Automotive Engineering, Wuhan University of Technology (WUT) in Jun. 2018. To chase his AI dream, he took the National Postgraduate Entrance Examination and obtained 1st place in School of Computer Science and Technology, University of Science and Technology of China (USTC), in Mar. 2018. He also obtained his master's degree from here in Jun. 2021. His research interests are mainly on image/video/3D understanding. 
\end{IEEEbiography}

\begin{IEEEbiography}[{\includegraphics[width=1in,height=1.25in,clip,keepaspectratio]{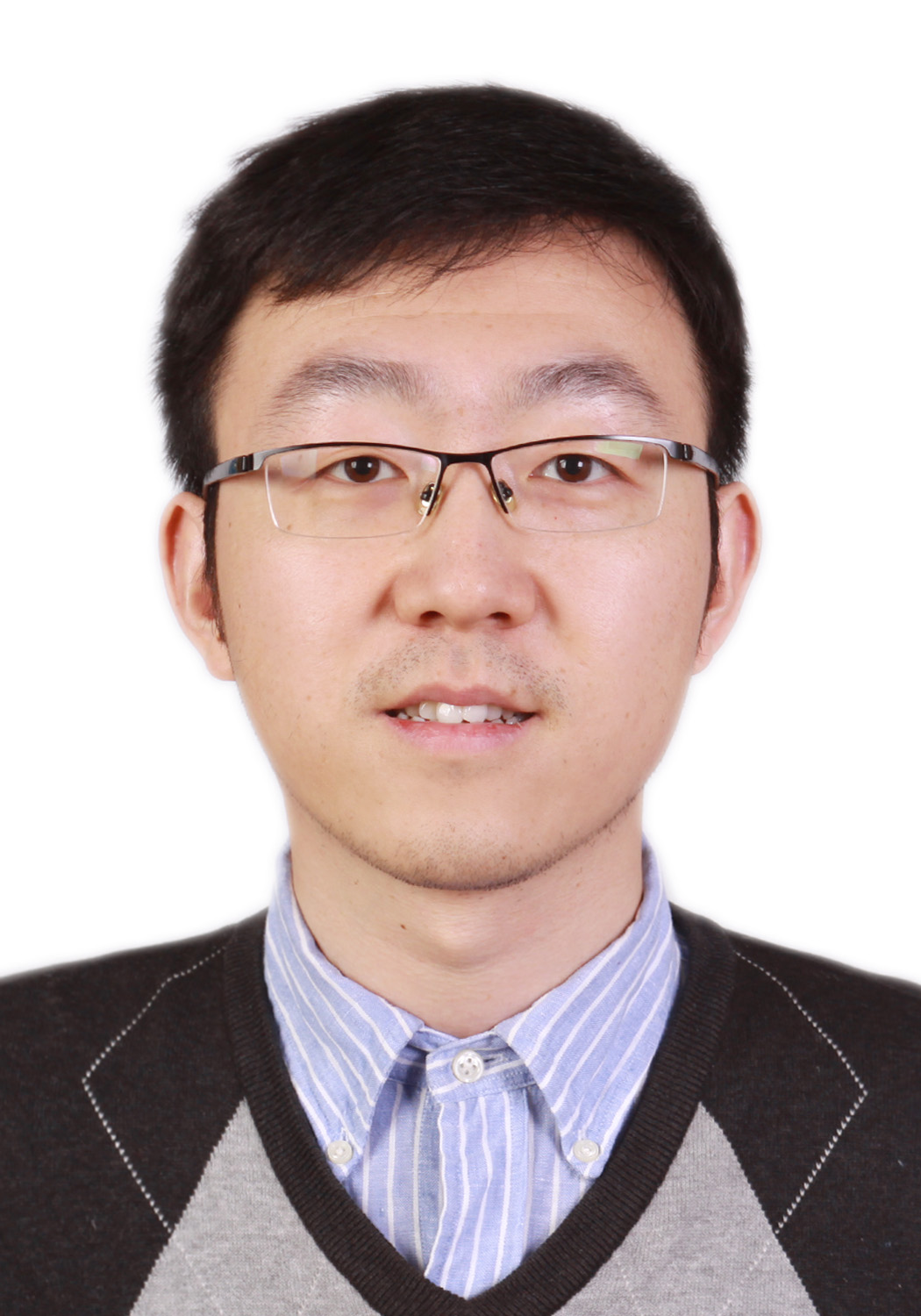}}]{Dong Liu}
  (SM'19) received the B.S. and Ph.D. degrees in electrical engineering from the University of Science and Technology of China (USTC), Hefei, China, in 2004 and 2009, respectively. He was a Member of Research Staff with Nokia Research Center, Beijing, China, from 2009 to 2012. He joined USTC in 2012 and became a Professor in 2020. His research interests include image and video processing, coding, analysis, and data mining. He has authored or co-authored more than 100 papers in international journals and conferences. He has 19 granted patents. He received the 2009 IEEE TCSVT Best Paper Award and the VCIP 2016 Best 10\% Paper Award. He is a Senior Member of IEEE, CCF, and CSIG, and an elected member of MSA-TC of IEEE CAS Society. He serves or had served as the Chair of IEEE Future Video Coding Study Group (FVC-SG), a Publicity Co-Chair for ICME 2021, a Registration Co-Chair for ICME 2019, and a Symposium Co-Chair for WCSP 2014.
\end{IEEEbiography}

\begin{IEEEbiography}[{\includegraphics[width=1in,height=1.25in,clip,keepaspectratio]{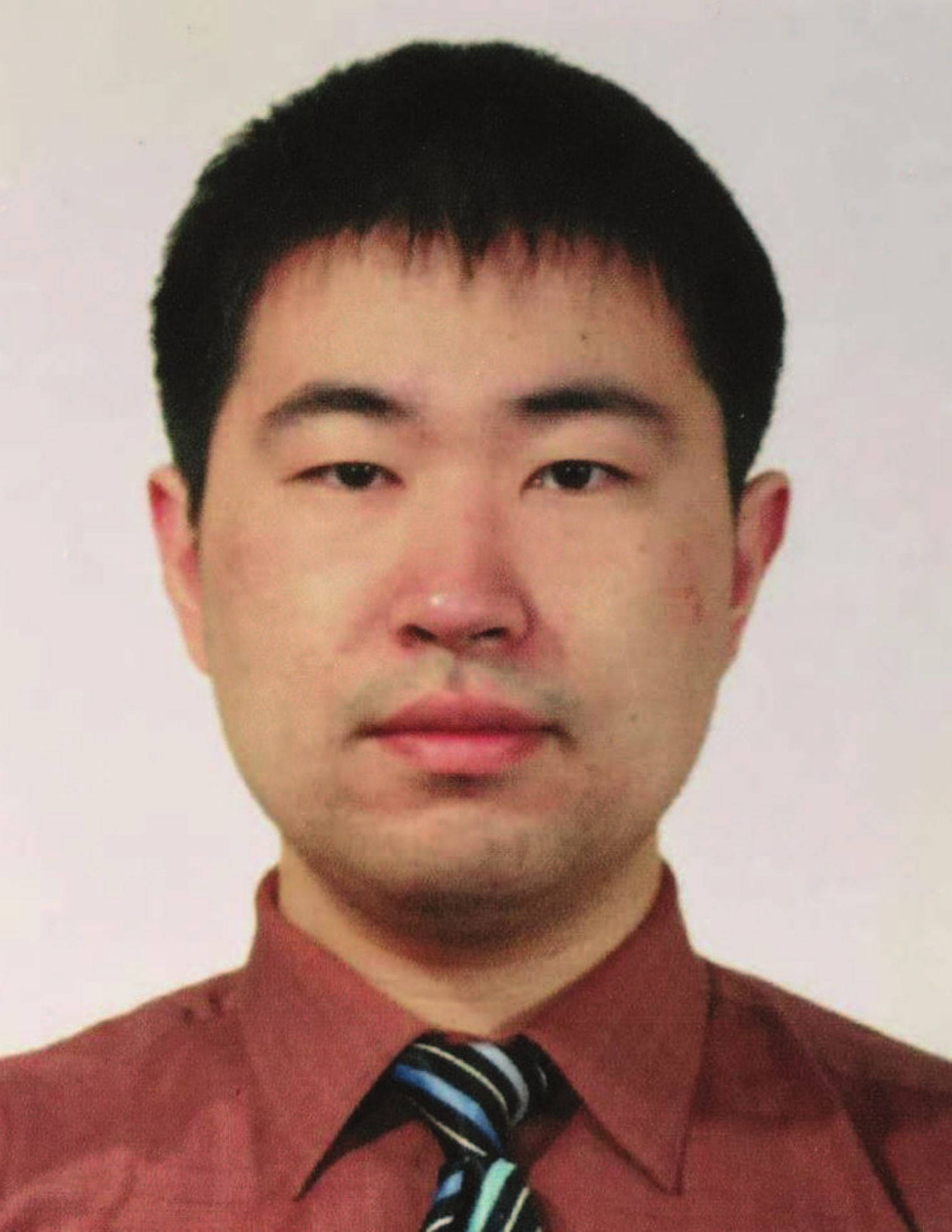}}]{Tong Xu}
  (M'17) is currently working as an Associate Professor of the Anhui Province Key Laboratory of Big Data Analysis and Application, University of Science and Technology of China (USTC), Hefei, China. He received the Ph.D. degree in USTC, 2016. He has authored more than 70 top-tier journal and conference papers in the fields of social network and management computing analysis, including TKDE, TMC, TMM, TOMM, KDD, SIGIR, WWW, AAAI, IJCAI, etc. Besides, he has served on numerous conferences, such as the track chair of IEEE ICKG 2020, the session chair of CCKS 2018 and SMP 2020, the co-organizers for workshops held in KDD 2018-2021 and SDM 2020-2021, and the program committee member of several international conferences like KDD, AAAI, WWW, IJCAI, etc. He was the recipient of the Best Paper Award of KSEM 2020.\par 
\end{IEEEbiography}

\begin{IEEEbiography}[{\includegraphics[width=1in,height=1.25in,clip,keepaspectratio]{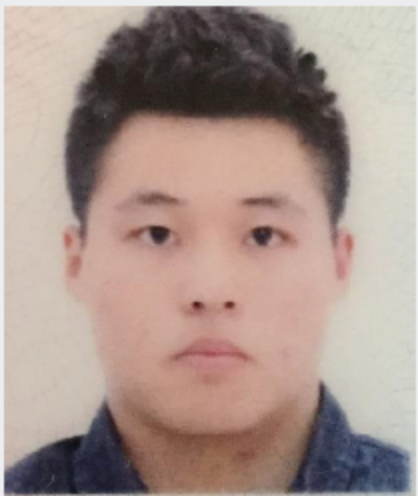}}]{Shiwei Wu}
  received the B.E. degree in from the Xiamen University (XMU), Xiamen, China, in 2018. He is currently working towards the Ph.D. degree with USTC, Hefei, China. He is working with Key Laboratory of Big Data Analysis and Application. His major research interests include video understanding and multimedia.
\end{IEEEbiography}

\begin{IEEEbiography}[{\includegraphics[width=1in,height=1.25in,clip,keepaspectratio]{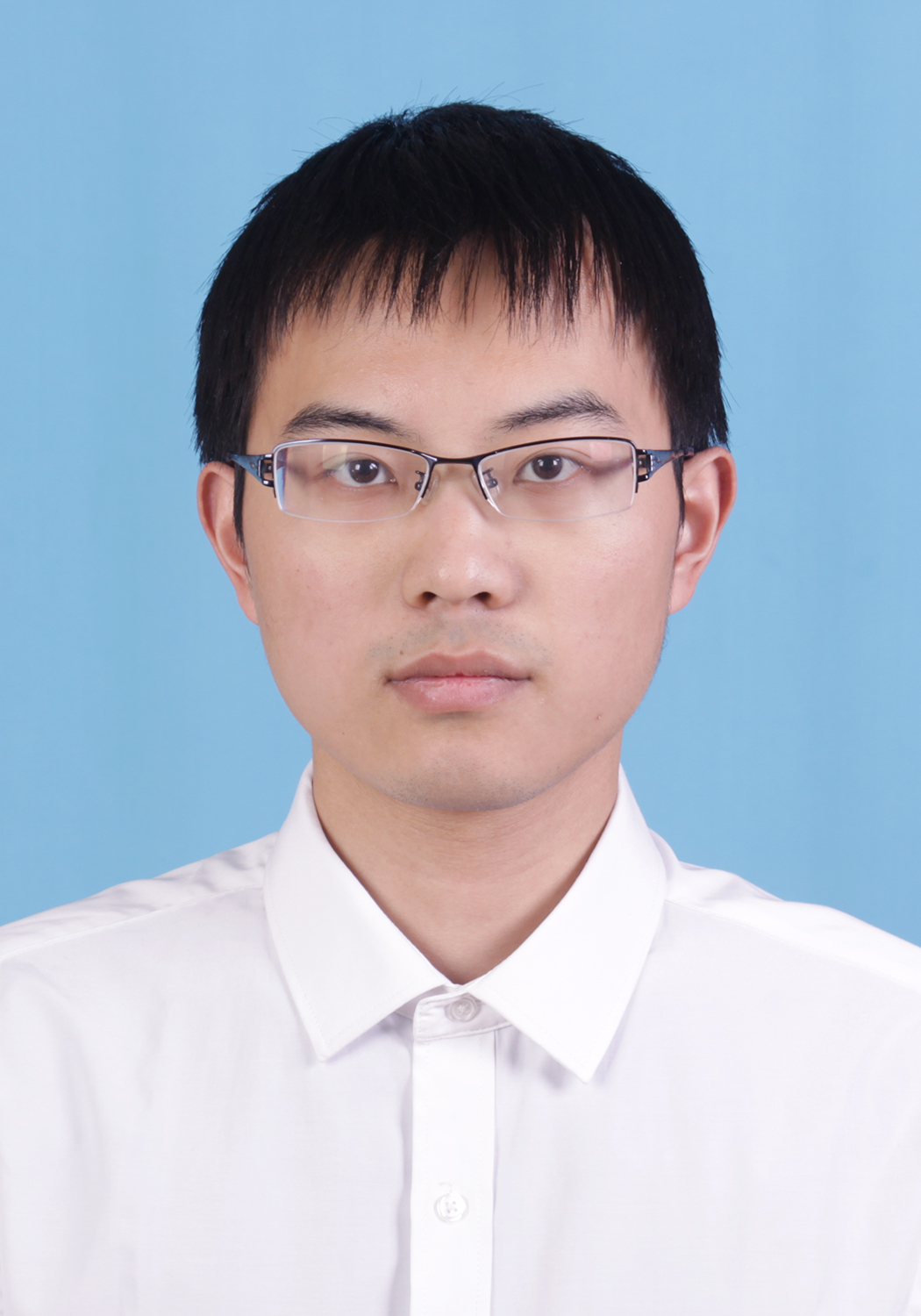}}]{Yifei Cheng}
  received the B.S. degree from the University of Science and Technology of China (USTC), Hefei, China, in 2015. He is currently working towards the Ph.D. degree with USTC, Hefei, China. He is working with Key Laboratory of Big Data Analysis and Application. His major research interests include optimization and federated learning.
\end{IEEEbiography}

\begin{IEEEbiography}[{\includegraphics[width=1in,height=1.25in,clip,keepaspectratio]{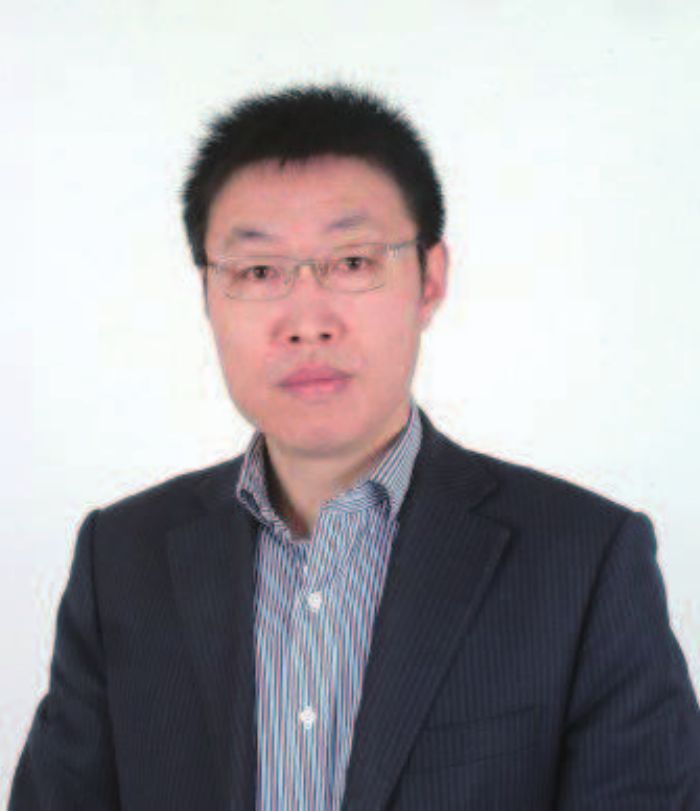}}]{Enhong Chen}
  (SM'07) is a Professor, executive dean of School of Data Science and vice dean of School of Computer Science of University of Science and Technology of China(USTC), CCF Fellow, IEEE Senior Member (Since 2007), winner of the National Science Fund for Distinguished Young Scholars (in 2013), scientific and technological innovation leading talent of 'Ten Thousand Talent Program'(in 2017) and member of the Decision Advisory Committee of Shanghai (Since June, 2018). I am also the vice director of the National Engineer Laboratory for Speech and Language Information Processing, the director of Anhui Province Key Laboratory of Big Data Analysis and Application, and the chairman of Anhui Province Big Data Industry Alliance. I received my B.Sc degree from Anhui University in 1989, Master degree from Hefei University of Technology in 1992 and Ph.D. degree in computer science from USTC in 1996. My current research interests are data mining and machine learning, especially social network analysis and recommender systems. I have published more than 200 papers on many journals and conferences, including international journals such as IEEE Trans, ACM Trans, and important data mining conferences, such as KDD, ICDM, NIPS. My research is supported by the National Natural Science Foundation of China, National High Technology Research and Development Program 863 of China, etc. I won the Best Application Paper Award on KDD2008 and Best Research Paper Award on ICDM2011.
\end{IEEEbiography}

\end{document}